\documentclass[10pt,twocolumn,letterpaper]{article}

\usepackage[pagenumbers]{cvpr} % To force page numbers, e.g. for an arXiv version

\usepackage[dvipsnames, table]{xcolor}

\definecolor{cvprblue}{rgb}{0.21,0.49,0.74}
\usepackage[pagebackref,breaklinks,colorlinks,citecolor=cvprblue]{hyperref}

\usepackage{times}
\usepackage{epsfig}
\usepackage{amsmath}

\usepackage{array}
\usepackage{subcaption}
\usepackage{multirow,multicol}
\usepackage[flushleft]{threeparttable}
\usepackage{comment}
\usepackage{algorithmic}
\usepackage{booktabs}
\usepackage{bbm, dsfont}
\usepackage[font=small,labelfont=bf]{caption}

\usepackage{amssymb}% http://ctan.org/pkg/amssymb
\usepackage{pifont}% http://ctan.org/pkg/pifont

\newcommand{\cmark}{\text{\ding{51}}}%
\newcommand{\xmark}{\text{\ding{55}}}%
\newcommand{\cmarkColor}{\cellcolor{green!15}\text{\ding{51}}}%
\newcommand{\xmarkColor}{\cellcolor{red!15}\text{\ding{55}}}%

\newcommand{\tablestyle}[2]{\setlength{\tabcolsep}{#1}\renewcommand{\arraystretch}{#2}\centering\footnotesize}

\newcolumntype{x}[1]{>{\centering\arraybackslash}p{#1pt}}
\newcommand{\app}{\raise.17ex\hbox{$\scriptstyle\sim$}}

\newlength\savewidth

\makeatletter\renewcommand\paragraph{\@startsection{paragraph}{4}{\z@}
  {.5em \@plus1ex \@minus.2ex}{-.5em}{\normalfont\normalsize\bfseries}}\makeatother

\def\fig#1{Fig.~\ref{fig:#1}}

\def\tablecite#1#{%
  \def\pretablecite{#1}%
  \tableciteaux}
\def\tableciteaux#1{%
  \textsuperscript{\expandafter\originalcite\pretablecite{#1}}%
}

\usepackage{graphicx}
\usepackage{enumitem}
\usepackage{wrapfig}
\usepackage{lipsum}
\usepackage{soul}

\newcolumntype{H}{>{\setbox0=\hbox\bgroup}c<{\egroup}@{}}
\newcolumntype{a}{>{\columncolor{Gray}}c}
\DeclareRobustCommand{\colorrowtext}[0]{{\sethlcolor{Gray}\hl{gray}}}
\usepackage{tabu}
\usepackage{nicematrix}

\definecolor{ForestGreen}{rgb}{0.13, 0.55, 0.13}
\definecolor{Green}{rgb}{0.0, 0.5, 0.0}
\definecolor{green(munsell)}{rgb}{0.0, 0.66, 0.47}
\definecolor{green(ryb)}{rgb}{0.4, 0.69, 0.2}
\definecolor{green(pigment)}{rgb}{0.0, 0.65, 0.31}
\definecolor{citecolor}{HTML}{0071bc}
\definecolor{GrayXMark}{gray}{0.7}

\usepackage{tabularx}
\usepackage[export]{adjustbox}

\newcommand{\ours}{VideoCutLER\xspace}
\newcommand{\vidsy}{ImageCut2Video\xspace}

\newcommand{\imnet}{ImageNet\xspace}

\definecolor{ForestGreen}{rgb}{0.13, 0.55, 0.13}
\definecolor{Green}{rgb}{0.0, 0.5, 0.0}
\definecolor{green(munsell)}{rgb}{0.0, 0.66, 0.47}
\definecolor{green(ryb)}{rgb}{0.4, 0.69, 0.2}
\definecolor{green(pigment)}{rgb}{0.0, 0.65, 0.31}

\newcommand{\plus}[1]{\small\bf\textcolor{Green}{#1}}

\newcolumntype{x}[1]{>{\centering\let\newline\\\arraybackslash\hspace{0pt}}p{#1}}
\definecolor{Gray}{gray}{0.9}
\newcommand{\tc}[1]{\textcolor{gray}{#1}}

\usepackage{makecell}

\usepackage[capitalize]{cleveref}
\crefname{section}{Sec.}{Secs.}
\Crefname{section}{Section}{Sections}
\Crefname{table}{Table}{Tables}
\crefname{table}{Table}{Tabs.}

\def\paperID{*****} % *** Enter the Paper ID here
\def\confName{CVPR}
\def\confYear{2024}

\title{VideoCutLER: Surprisingly Simple Unsupervised Video Instance Segmentation}

\author{
  Xudong Wang \quad \quad
  Ishan Misra \quad \quad
  Ziyun Zeng \quad \quad
  Rohit Girdhar \quad \quad
  Trevor Darrell \\
  UC Berkeley \quad \quad Meta AI \\
  \small{Code:} \href{https://github.com/facebookresearch/CutLER}{\small{https://github.com/facebookresearch/CutLER}}
}

\begin{document}
\twocolumn[{%
  \renewcommand\twocolumn[1][]{#1}%
  \maketitle
    \vspace{-19pt}
    \captionsetup{type=figure}
    \centering
    \includegraphics[width=1.0\textwidth]{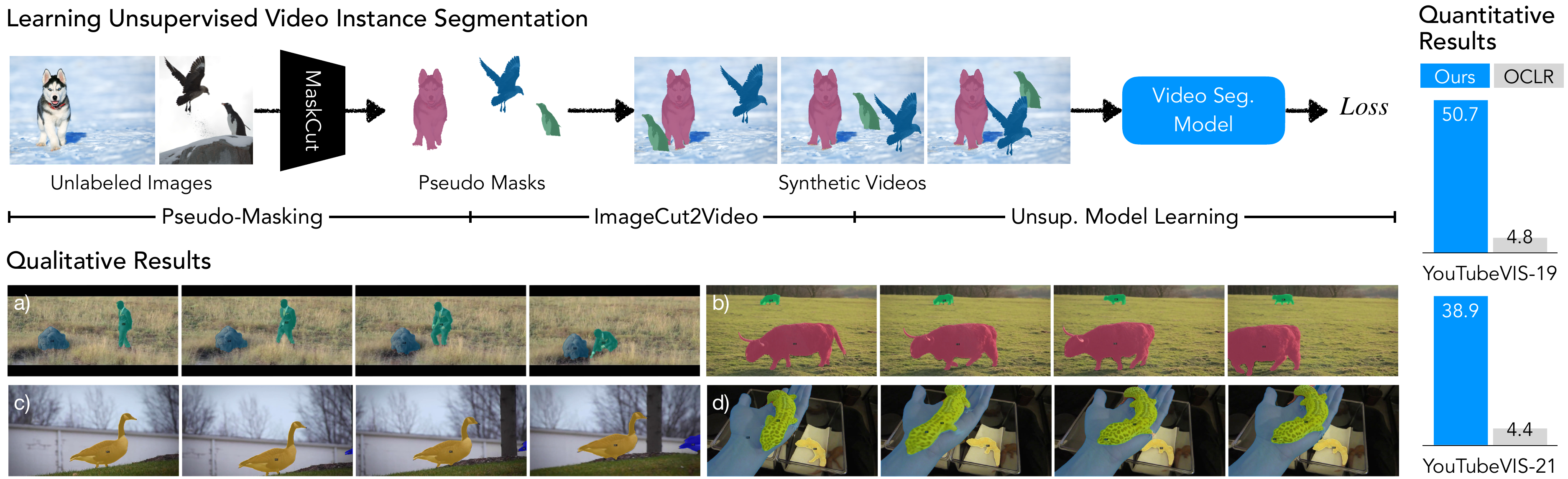}
    \vspace{-15pt}
    \caption{
      \ours is a simple unsupervised video instance segmentation method (UnVIS). We show the first competitive unsupervised results on the challenging YouTubeVIS benchmark.
      Moreover, unlike most prior approaches, we demonstrate that UnVIS models can be learned without relying on natural videos and optical flow estimates. 
      \textbf{\textit{Row 1:}}
      We propose \textbf{VideoCutLER}, a simple cut-synthesis-and-learn pipeline that involves three main steps.
      Firstly, we generate pseudo-masks for multiple objects in an image using MaskCut~\cite{wang2023cut}.
      Then, we convert a random pair of images in the minibatch into a video with corresponding pseudo mask trajectories using ImageCut2Video. 
      Finally, we train an unsupervised video instance segmentation model using these mask trajectories.
      \textbf{\textit{Row 2:}}
      Despite being trained only on unlabeled images, at inference time \ours can be directly applied to unseen videos and can segment and track multiple instances across time (Fig.~\ref{fig:teaser}a), even for small objects (Fig.~\ref{fig:teaser}b), objects that are absent in specific frames (Fig.~\ref{fig:teaser}c), and instances with high overlap (Fig.~\ref{fig:teaser}d). 
      \textbf{\textit{Column~2:}}
      Our method surpasses the previous SOTA method OCLR~\cite{xie2022segmenting} by a factor of 10 in terms of class-agnostic AP$^{\text{video}}_{50}$.
    }
    \label{fig:teaser}
    \vspace{16pt}
}]

% \maketitle
% Remove page # from the first page of camera-ready.
% \ificcvfinal\thispagestyle{empty}\fi

\begin{abstract}
    %Unsupervised video instance segmentation is a crucial computer vision task enabling a deep understanding of video contents without human annotations.
    Existing approaches to unsupervised video instance segmentation typically rely on motion estimates and experience difficulties tracking small or divergent motions.
    We present \ours, a simple method for unsupervised multi-instance video segmentation without using motion-based learning signals like optical flow or training on natural videos.
    Our key insight is that using high-quality pseudo masks and a simple video synthesis method for model training is surprisingly sufficient to enable the resulting video model to effectively segment and track multiple instances across video frames.
    We show the first competitive unsupervised learning results on the challenging YouTubeVIS-2019 benchmark, achieving 50.7\% AP$^{\text{video}}_{50}$, surpassing the previous state-of-the-art by a large margin. 
    \ours can also serve as a strong pretrained model for supervised video instance segmentation tasks, exceeding DINO by 15.9\% on YouTubeVIS-2019 in terms of AP$^{\text{video}}$.
    % We will release the code.
    %We will open-source the code. \tjd{try really hard to have abstract fit in first col. first and last sentences can be cut IMHO: they are appropriate for intro not abs.}
    % Benchmarked on YouTubeVIS-2019 and YouTubeVIS-2021, \ours sets a new state-of-the-art performance with 47.8\% AP$_{50}$ and 36.8\% AP$_{50}$, respectively, surpassing the previous SOTA by 43\% (47.8\% vs. 4.8\%) and 32.4\% (36.8\% vs. 4.4\%), respectively.
    % To our best knowledge, this is the first work that shows the possibility of getting a SOTA video segmentation model without leveraging optical flow or motion estimation modules.
    % of solely training on ImageNet-1K images for unsupervised video instance segmentation.
\end{abstract}

% Need to add something that is affirmtive. 
% The first sentence describes the motivations. Second drawbacks of prior works.
% Third describe 

% Suprisingly Simple Synthetic Video Generation, to provide the first succefully result on YoutubeVIS-2019. 

% Go affirmtive!!

% Cut-based segmentation and a very simple synthetic 

% Adding a more powerful motion estimates may further improve.

% Explicit motion estimates, such as optical flow. 

% Typically not typically highly 

% Cut a lot of things!!!!

% Don't use numerations. 

% Making three points:
% 1) Insights: Prior work is over replying on 
% 2) Method: 
% 3) Results: We figure out a simple way to 
\def\figVSOCLR#1{
    \captionsetup[sub]{font=small}
    \begin{figure}[#1]
      \centering
      \includegraphics[width=1.0\linewidth]{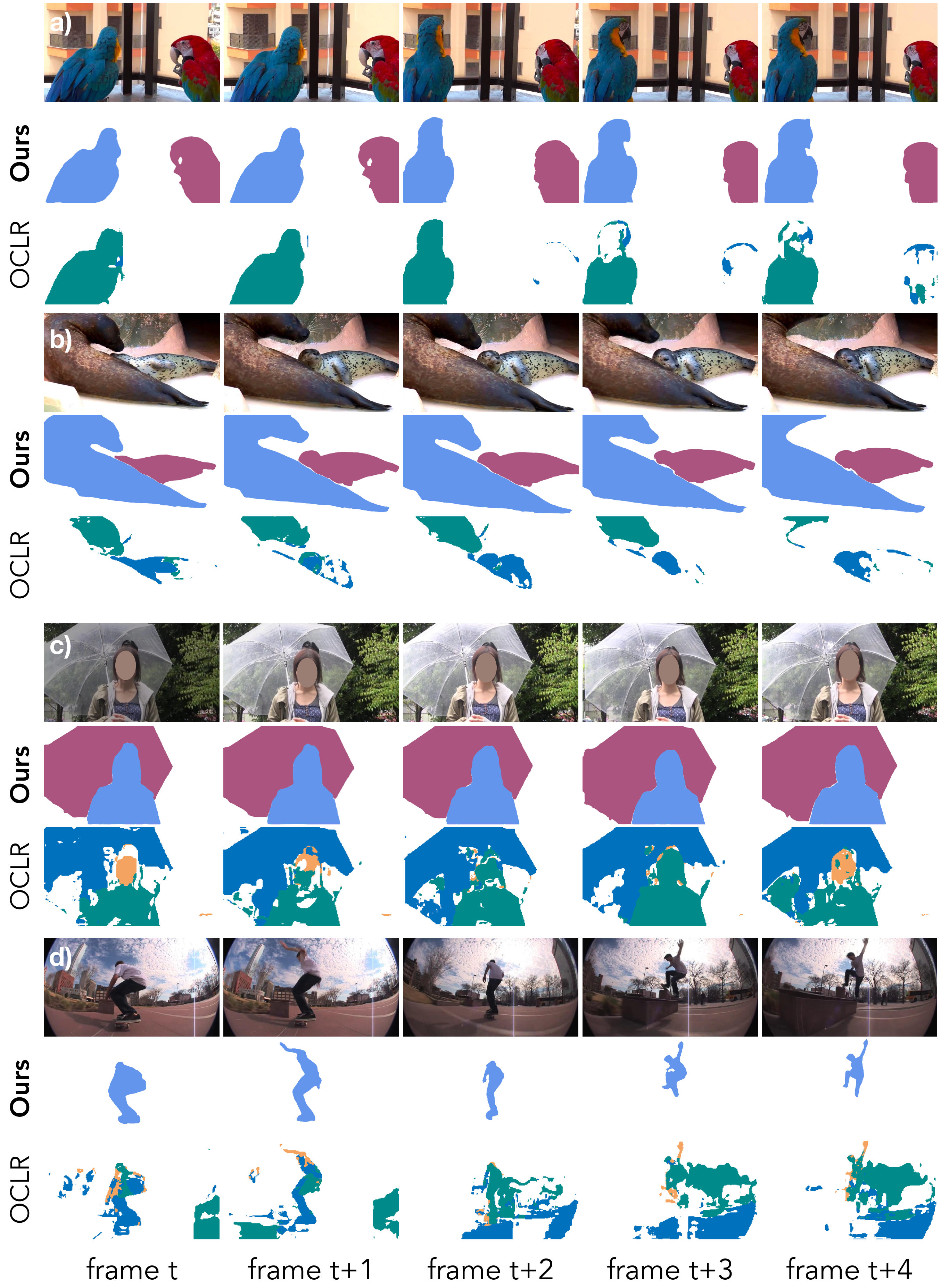}
      % \vspace{-6pt}
      \caption{
        \textbf{Challenges encountered by the previous state-of-the-art} OCLR: Within the framework of OCLR~\cite{xie2022segmenting}, a method that heavily relies on optical flows as model inputs, several distinct failure cases emerge. These include situations where the method struggles to accurately segment both moving and static objects (as demonstrated in Fig.~\ref{fig:vs-oclr}a), struggles to effectively track non-rigid objects as a coherent unit (Fig.~\ref{fig:vs-oclr}b), encounters difficulties in distinguishing overlapping instances (Fig.~\ref{fig:vs-oclr}c), and fails to maintain consistent predictions under varying illumination conditions (Fig.~\ref{fig:vs-oclr}d).
        Nonetheless, many of these challenges can be effectively addressed through the application of our proposed approach, \ours, without being reliant on the optical estimations used by various prior works~\cite{yang2021self,xie2022segmenting}. 
        We present qualitative comparisons using the YouTubeVIS dataset~\cite{yang2019video}.
      }
      \label{fig:vs-oclr}
    \end{figure}
}

\section{Introduction}
% Video segmentation is a fundamental task in computer vision that involves partitioning a video sequence into meaningful segments corresponding to distinct objects or regions of interest~\cite{wang2021survey}. This task plays a crucial role in various practical applications such as video surveillance, autonomous driving, and video editing, as it enables a deep understanding of the video content and facilitates more advanced processing and analysis.
% However, it is costly to collect such datasets and the corresponding labels. Hence, there is a pressing need to devise an unsupervised learning approach for video instance segmentation to comprehend video content comprehensively.

Video instance segmentation is vital for many computer vision applications, \eg video surveillance, autonomous driving, and video editing, yet labeled videos are costly to obtain. 
Hence, there is a pressing need to devise an unsupervised video instance segmentation approach that can comprehend video content comprehensively and operate in general domains without labels.

\figVSOCLR{t!}

Prior work in this area typically relies on an optical flow network as an off-the-shelf motion estimator~\cite{xie2022segmenting,teed2020raft,yang2021self}. 
Although optical flow can be informative in detecting pixel motion between frames, it is not always a reliable technique, particularly in the presence of occlusions, motion blur, complex motion patterns, changes in illuminations, \etc.
As a result, models that heavily rely on optical flow estimations may fail in several common scenarios. For example, stationary or slowly moving objects may have flow estimates similar to the background, causing them to be omitted in the segmentation process (\eg, the parrot with negligible motion is missed in \cref{fig:vs-oclr}a). 
Similarly, non-rigid objects with non-consistent motions for several parts have varying optical flows, leading to a failure in segmenting all parts cohesively as a unit if object motion is presumed constant (\cref{fig:vs-oclr}b).
Also, objects with similar motion patterns and high overlap are complex for optic flow methods to accurately distinguish between them, especially in boundary regions (\cref{fig:vs-oclr}c).
Finally, objects with illumination changes across frames can cause optical-flow based models to produce non-consistent and blurred segmentation masks (\cref{fig:vs-oclr}d).  
Given the limitations above, we advocate for unsupervised video segmentation models which do not depend on optical flow estimates. 
%
%We find that a frustrating simple video synthesis method that does not rely on explicit motion estimates or natural videos is surprisingly effective in learning an unsupervised multi-instance video segmentation model and addressing aforementioned limitations, as illustrated in \cref{fig:vs-oclr}.
%
We propose a method to train a video segmentation model by generating simple synthetic videos from individual images, without relying on explicit motion estimates or requiring labeled natural videos.

% \fixme{acknowledge that video synthesis is not new, and explain the key differences.}
%
% \fixme{explain that motion is not that important, the appearance similarity tracking is an important and more powerful cue.}
%
% Given unlabeled images, training a model to perform simultaneous instance segmentation and tracking on videos poses two main challenges: 1) extracting precise object proposals with accurate boundaries from the unlabeled images, and 2) enabling the model to associate object proposals across frames after training on these unlabeled images and corresponding pseudo masks.
% Pseudo-masking can be acomplished by CutLER~\cite{wang2023cut}, which is the state-of-the-art unsupervised image instance segmentation model. 
% This work, \ours, extends CutLER's capability to the video domain, aiming to perform object segmentation with temporal consistency across video frames.
%
Our method, \textbf{\ours}, is an unsupervised \textbf{Video} instance segmentation model that employs a \textbf{Cu}t-syn\textbf{t}hesis-and-\textbf{LE}a\textbf{R}n pipeline (\cref{fig:teaser}).
First, given unlabeled images, we extract pseudo-masks for multiple objects in an image using MaskCut~\cite{wang2023cut}, leveraging a self-supervised DINO~\cite{caron2021emerging} and a spectral clustering method Normalized Cuts~\cite{shi2000normalized} (details in \cref{sec:background}).
% We use a self-supervised DINO model's features~\cite{caron2021emerging} to create a patch-wise similarity matrix for the image. We then iteratively apply normalized cuts~\cite{shi2000normalized} on a masked similarity matrix to extract multiple pseudo-masks.
%
Second, given unlabeled images and their pseudo-masks in a minibatch, we propose \vidsy, a surprisingly simple video synthesis scheme that generates a video from those with corresponding pseudo mask trajectories (details in \cref{sec:imgcut2vid}).
Finally, those mask trajectories are used to train a video instance segmentation model, aiming to perform object segmentation with temporal consistency across video frames (details in \cref{sec:model_learning}).
%\tjd{rewrite the above para without enumeration or italics. Use normal prose and "First," "Second," "Finally,", etc.}

We utilize VideoMask2Former~\cite{cheng2021mask2former} as our video instance segmentation model, which operates by attending to the 3D spatiotemporal features of our synthetic videos and generating 3D volume predictions of pseudo-mask trajectories using shared queries across frames.
The shared queries across frames enable the model to segment and track object instances based on their appearance (feature) similarities. 

Despite being learned from only unlabeled images (and the temporally simple synthetic video sequences we construct from them), \ours succeeds at multi-instance video segmentation, achieving a new state-of-the-art (SOTA) performance of 50.7\% AP$^{\text{video}}_{50}$ on YouTubeVIS-2019. 
This result surpasses the previous SOTA~\cite{xie2022segmenting} by substantial margins of 45.9\% (50.7\% vs. 4.8\%).
This result also considerably narrows the performance gap between supervised and unsupervised learning, reducing it from 29.1\% to 11.0\% in terms of the AP$^{\text{video}}_{50}$.

Moreover, most prior works on self-supervised representation learning~\cite{he2020momentum,chen2020simple,grill2020bootstrap,wang2021unsupervised,caron2021emerging} are limited to providing initializations only for the model backbones, with the remaining layers being randomly initialized.
In contrast, our pretraining strategy takes a more comprehensive approach that allows all model weights to be pretrained, resulting in a stronger pretrained model better suited for supervised learning.
As a result, our method outperforms DINO's~\cite{caron2021emerging} AP$^{\text{video}}$ on YoutubeVIS-2019 by 15.9\%.

Our work makes the following contributions:
\textit{\textbf{Insights:}} 
We found that a simple video synthesis method yield surprisingly effective results for training unsupervised multi-instance video segmentation models. Importantly, this efficacy is achieved without the necessity of explicit motion estimates or the utilization of natural videos, a novel aspect that has not been previously demonstrated in the field.
\textit{\textbf{Methods:}} We propose a simple yet effective cut-synthesize-and-learn pipeline \ours for learning video instance segmentation models, given unlabeled images. 
\textit{\textbf{Results:}} Our method shows the first successfully results on challenging unsupervised multi-instance video segmentation benchmark YouTubeVIS, outperforming the previous SOTA model's AP$^{\text{video}}_{50}$ by a large margin.
\def\tabDistMethods#1{
    \begin{table}[#1]
    \centering
    \tablestyle{3pt}{1.2}
    \small
    \begin{tabular}{lccccc}
    \Xhline{0.8pt}
    & CRW & DINO & OCLR & Ours \\ 
    \hline
    Segment multiple objects & \cmarkColor & \xmarkColor &  \cmarkColor & \cmarkColor \\
    Track objects across frames & \cmarkColor & \xmarkColor & \cmarkColor & \cmarkColor \\
    No need for optical flow & \cmarkColor & \cmarkColor & \xmarkColor & \cmarkColor \\
    % zero-shot segmentation & \cmarkColor & \cmarkColor & \cmarkColor & \cmarkColor \\\hline
    \begin{tabular}[c]{@{}l@{}}No 1st-frame ground-truth \end{tabular} & \xmarkColor & \xmarkColor & \cmarkColor & \cmarkColor \\
    \begin{tabular}[c]{@{}l@{}}No human labels at any stage\end{tabular} & \xmarkColor & \xmarkColor & \phantom{1}\cmarkColor$^{\dagger}$ & \cmarkColor \\
    \begin{tabular}[c]{@{}l@{}}Pretrained model for sup. learning \end{tabular} & \xmarkColor & \xmarkColor & \xmarkColor & \cmarkColor \\
    % \begin{tabular}[c]{@{}l@{}}compatible with various \\ segmentation architectures \end{tabular} & \xmarkColor & \graydash & \cmarkColor & \xmarkColor & \cmarkColor \\
    \Xhline{0.8pt}
    \end{tabular}
    % \vspace{-2pt}
    \caption{
    We compare previous methods on unsupervised instance segmentation, including CRW~\cite{jabri2020space}, DINO~\cite{caron2021emerging}, and OCLR~\cite{xie2022segmenting}, with our \ours in term of key properties. 
    Our \ours is the only approach that fulfills all these desired properties.
    $\dagger$: The optical flow estimator OCLR employs (RAFT~\cite{teed2020raft}) is pretrained on both synthetic data and human-annotated data like KITTI-2015~\cite{keuper2015motion} and HD1K~\cite{kondermann2016hci}.
    }
    \label{tab:distMethods}
    \end{table}
}

\section{Related Work}

\noindent \textbf{Unsupervised video instance segmentation} (VIS) requires not only separating and tracking the main moving foreground objects from the background, but also differentiating between different instances, without any human annotations~\cite{wang2021survey}. Previous works~\cite{hu2018unsupervised,yang2019unsupervised,yang2021self,wang2022tokencut,lee2023unsupervised} on unsupervised video segmentation has primarily centered on \textbf{\textit{unsupervised video object segmentation}} (VOS), aiming to detect all moving objects as the foreground and to generate a pixel-level binary segmentation mask, regardless of whether the scene contains a single instance or multiple instances.
Despite some works exploring \textbf{\textit{unsupervised video instance segmentation}} (VIS), many of these approaches have resorted to either utilizing first frame annotations~\cite{li2019joint,jabri2020space,caron2021emerging} to propagate label information throughout the video frames or leveraging supervised learning using large amounts of external labeled data~\cite{ventura2019rvos,yang2019videoseg,zhou2020matnet,luiten2020unovost}. 
Furthermore, prior studies typically utilized optical flow networks that were pretrained with human supervision using either synthetic data or labeled natural videos~\cite{yang2019unsupervised,ventura2019rvos,yang2021self,xie2022segmenting}.

The properties deemed necessary for an unsupervised learning method to excel in video instance segmentation tasks are presented and discussed in \cref{tab:distMethods}. Our proposed method, \ours, is the only approach that satisfies all these properties, making it an effective and promising solution for unsupervised video instance segmentation.

\tabDistMethods{t!}

\noindent \textbf{Unsupervised object discovery} aims to automatically discover and segment objects in an image in an unsupervised manner~\cite{wang2022tokencut,hu2018unsupervised,wang2022freesolo,wang2023cut}.
LOST~\cite{simeoni2021localizing} and TokenCut~\cite{wang2022tokencut} focus on salient object detection and segmentation via leveraging the patch features from a pretrained DINO~\cite{caron2021emerging} model. 
For multi-object discovery, FreeSOLO~\cite{wang2022freesolo} first generates object pseudo-masks for unlabeled images, then learns an unsupervised instance segmentation model using these pseudo-masks.
CutLER~\cite{wang2023cut} presents a straightforward cut-and-learn pipeline for unsupervised detection and segmentation of multiple instances. It has demonstrated promising results on more than eleven different benchmarks, covering a wide range of domains. 

In contrast to previous approaches, our unsupervised learning method focuses on simultaneously tracking objects in a video sequence while identifying correspondences between instances across multiple frames.

\noindent \textbf{Self-supervised representation learning} generates its own supervision signal by exploiting the implicit patterns or structures present in the input data~\cite{he2020momentum,caron2020unsupervised,caron2021emerging,he2022masked}. 
Unlike most previous self-supervised learning models, which still require fine-tuning on labeled data to be operative on complex computer vision tasks, such as detection and segmentation, \ours can tackle these complex, challenging tasks with purely unsupervised learning methods.
%%%%%%%%%%%%%%%%%%%%%%%%%%%%%%%%%%%%%%%%%%%%%%%%%%%%%%
\def\figVideoCutLER#1{
    \captionsetup[sub]{font=small}
    \begin{figure*}[#1]
      \centering
      \includegraphics[width=0.99\linewidth]{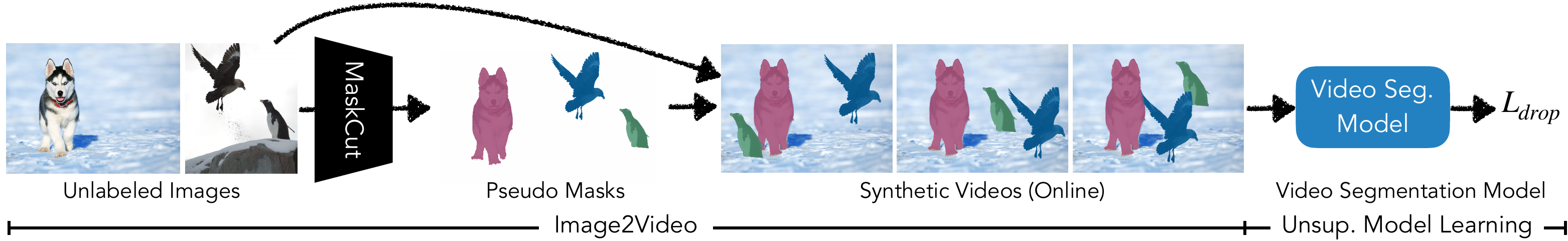}\vspace{-3pt}
      \caption{
      The \textbf{cut-synthesize-and-learn pipeline} of VideoCutLER consists of two main steps. First, we use MaskCut to generate pseudo-masks for multiple objects in an image. Next, we learn an unsupervised video instance segmentation model using Image2Video, which converts a pair of images in the minibatch into a video with corresponding pseudo mask trajectories.
      After training exclusively on ImageNet-1K \ours can discover and track multiple objects in a video without supervision.
      }
      \label{fig:videoculer}
    \end{figure*}
}

\def\tabUnsupYTVIS#1{
\begin{table*}[#1]
% \vspace{-16pt}
\tablestyle{1.9pt}{1.0}
\small
\begin{center}
\begin{tabular}{p{2.1cm}x{0.9cm}x{0.9cm}x{0.8cm}lccccccclccccccc}
\Xhline{0.8pt}
\multirow{2}{*}{Methods} & \multicolumn{3}{c}{{Training settings}} && \multicolumn{7}{c}{{YouTubeVIS-2019}} && \multicolumn{7}{c}{{YouTubeVIS-2021}} \\
\cline{2-4} \cline{6-12} \cline{14-20}
& flow & videos & sup. && \multicolumn{1}{c}{AP$_{50}$} & \multicolumn{1}{c}{AP$_{75}$} & \multicolumn{1}{c}{AP} & \multicolumn{1}{c}{AP$_S$} & \multicolumn{1}{c}{AP$_M$} & \multicolumn{1}{c}{AP$_L$} & \multicolumn{1}{c}{AR$_{10}$}
&& \multicolumn{1}{c}{AP$_{50}$} & \multicolumn{1}{c}{AP$_{75}$} & \multicolumn{1}{c}{AP} & \multicolumn{1}{c}{AP$_S$} & \multicolumn{1}{c}{AP$_M$} & \multicolumn{1}{c}{AP$_L$} & \multicolumn{1}{c}{AR$_{10}$} \\ [.1em]
\hline 
MotionGroup$^*$~\cite{yang2021self} & \cmark & \cmark & \xmark && \phantom{1}\phantom{1}1.3 & \phantom{1}\phantom{1}0.1 &  \phantom{1}\phantom{1}0.3 & \phantom{1}0.2 &  \phantom{1}\phantom{1}0.3 &  \phantom{1}\phantom{1}0.5 & \phantom{1}\phantom{1}1.7 && \phantom{1}\phantom{1}1.1 & \phantom{1}\phantom{1}0.1 & \phantom{1}\phantom{1}0.2 & \phantom{1}0.1 & \phantom{1}\phantom{1}0.2 & \phantom{1}\phantom{1}0.5 & \phantom{1}\phantom{1}1.5 \\
OCLR$^*$~\cite{xie2022segmenting} & \cmark & \cmark & \phantom{1}\xmark$^\dagger$ && \phantom{1}\phantom{1}4.8 & \phantom{1}\phantom{1}0.4 &  \phantom{1}\phantom{1}1.3 & \phantom{1}0.0 &  \phantom{1}\phantom{1}1.2 &  \phantom{1}\phantom{1}5.5 & \phantom{1}11.0 && \phantom{1}\phantom{1}4.4 & \phantom{1}\phantom{1}0.3 & \phantom{1}\phantom{1}1.2 & \phantom{1}0.1 & \phantom{1}\phantom{1}1.6 & \phantom{1}\phantom{1}7.1 & \phantom{1}\phantom{1}9.6 \\
% \hline
\rowcolor{Gray}
CutLER$^\ddagger$ & \xmark & \xmark & \xmark && \phantom{1}37.5 & \phantom{1}14.6 & \phantom{1}17.1 & \phantom{1}3.3 & \phantom{1}13.9 & \phantom{1}27.6 & \phantom{1}30.4 && 
\phantom{1}29.2 & \phantom{1}10.4 & \phantom{1}12.8 & \phantom{1}3.1 & \phantom{1}12.8 & \phantom{1}27.8 & \phantom{1}22.6 \\
\rowcolor{Gray}
\ours & \xmark & \phantom{1}\cmark$^{\divideontimes}$ & \xmark && \phantom{1}50.7 & \phantom{1}24.2 & \phantom{1}26.0 & \phantom{1}5.6 & \phantom{1}20.9 & \phantom{1}37.9 & \phantom{1}42.4 && \phantom{1}38.9 & \phantom{1}19.0 & \phantom{1}17.1 & \phantom{1}5.3 & \phantom{1}18.3 & \phantom{1}37.5 & \phantom{1}31.3 \\
\rowcolor{Gray} 
\textit{vs. prev. SOTA} &&&&& \plus{+12.8}  & \phantom{1}\plus{+9.6} & \phantom{1}\plus{+8.9} & \plus{+2.3} & \phantom{1}\plus{+7.0} & \plus{+10.3} & \plus{+12.0} && \phantom{1}\plus{+9.7} & \phantom{1}\plus{+8.6} & \phantom{1}\plus{+4.3} & \plus{+2.2} & \phantom{1}\plus{+5.5} & \phantom{1}\plus{+9.7} & \phantom{1}\plus{+8.7} \\
\Xhline{0.8pt}
% \shline
\end{tabular}
\end{center}
\vspace{-10pt}
\caption{\textbf{Zero-shot unsupervised multi-instance video segmentation} on YouTubeVIS-2019 and YouTubeVIS-2021. We report the instance segmentation metrics (AP and AR) and training settings.
$^*$: reproduced MotionGroup~\cite{yang2021self} and OCLR~\cite{xie2022segmenting} results with the official code and checkpoints. 
$^\dagger$: the optical flow estimator OCLR employs (RAFT~\cite{teed2020raft}) is pretrained on both synthetic data~\cite{dosovitskiy2015flownet,Butler:ECCV:2012} and human-annotated data, such as KITTI-2015~\cite{keuper2015motion} and HD1K~\cite{kondermann2016hci}.
$^{\ddagger}$: We train a CutLER~\cite{wang2023cut} model with Mask2Former as a detector on \imnet-1K, following CutLER's official training recipe, and use it as a strong baseline.
% VideoCutLER is solely trained on unlabeled ImageNet-1K. 
$^{\divideontimes}$: \ours is trained on synthetic videos generated using \imnet. 
Sup and flow denote human supervision and optical flow information, respectively. 
We evaluate results on YouTubeVIS's \texttt{train} splits in a class-agnostic manner (note: we never train on YouTubeVIS).
}
% \vspace{-16pt}
\label{tab:zero-shot-ytvis}
\end{table*}
}

% \figVideoCutLER{h!}
%%%%%%%%%%%%%%%%%%%%%%%%%%%%%%%%%%%%%%%%%%%%%%%%%%%%%%
\section{\ours}
\label{sec:method}
This section presents VideoCutLER, a simple cut-synthesis-and-learn pipeline consisting of three main steps. 
First, we generate pseudo-masks for multiple objects in an image using MaskCut (\cref{sec:background}).
Next, we convert a random pair of images in the minibatch into a synthetic video with corresponding pseudo mask trajectories using ImageCut2Video (\cref{sec:imgcut2vid}). 
Finally, we train an unsupervised video instance segmentation (VIS) model using these mask trajectories. As the model inputs do not contain explicit motion estimates, it learns to track objects based on their appearance similarity (\cref{sec:model_learning}). 
We will provide further details on each step in the following sections.

\subsection{Single-image unsupervised segmentation}
\label{sec:background}
We employ the MaskCut method, introduced in the CutLER~\cite{wang2023cut} method. MaskCut is an efficient spectral clustering approach for unsupervised image instance segmentation and object detection and can discover multiple object masks in a single image without human supervision. 
MaskCut builds upon a self-supervised DINO model~\cite{carion2020end} with a backbone of ViT~\cite{dosovitskiy2020image} and a cut-based clustering method Normalized Cuts (NCut)~\cite{shi2000normalized}. 
MaskCut first generates a patch-wise affinity matrix $W_{ij}\!=\!\frac{K_i K_j}{\|K_i\|_2 \|K_j\|_2}$ using the `key' features $K_i$ for patch $i$ from DINO's last attention layer. 
Subsequently, the NCut algorithm~\cite{shi2000normalized} is employed on the affinity matrix by solving a generalized eigenvalue problem
\begin{align}
    (D - W)x = \lambda Dx
\end{align}
where $D$ is a diagonal matrix with $d(i)=\sum_jW_{ij}$ and $x$ is the eigenvector that corresponds to the second smallest eigenvalue $\lambda$. 
Then, the foreground masks ${M}^{s}$ can be extracted via bi-partitioning $x$, which segments a single object within the image. 
To segment more than one instance, MaskCut employs an iterative process that involves masking out the values in the affinity matrix using the extracted foreground mask
\begin{align}
  W^{t}_{ij}\!=\!\frac{(K_i\prod_{s=1}^{t}{M}^s_{ij})(K_j\prod_{s=1}^{t}{M}^s_{ij})}{\|K_i\|_2\|K_j\|_2}
  \label{eqn:update-graph}
\end{align}
and repeating the NCut algorithm. MaskCut repeats this process $t$ times and sets $t\!=\!3$ by default.

% \textit{\textbf2) Unsupervised model learning.} 
% CutLER utilizes pseudo-masks generated by MaskCut to train a detector on ImageNet-1K. 
% To overcome the limitation of MaskCut in locating small objects, CutLER employs a loss-dropping strategy during training that encourages the detector to explore regions that MaskCut may overlook.
% Specifically, following~\cite{wang2023cut}, the loss for each predicted region $r_i$ is dropped (set as 0) if its maximum intersection over union (IoU) with any of the `pseudo-masks' is below a threshold $\tau^{\text{IoU}}$:
% \begin{align}
%   \mathcal{L}_{\text{drop}}(r_i) = (1-\mathbbm{1}(\text{IoU}_i^{\text{max}} \le \tau^{\text{IoU}}))\mathcal{L}_{\text{vanilla}}(r_i)
% \end{align}
% \noindent $\text{IoU}_i^{\text{max}}$ represents the maximum IoU of region $r_i$ with all pseudo-masks, and $\mathcal{L}{_\text{vanilla}}$ denotes the vanilla loss function of the video segmentation models.

% \textit{\textbf3) Self-training (optional). {CutLER uses a self-training strategy to improve the performance further and fine-tune the detector on the model's predictions. 

% While CutLER has demonstrated impressive performance in image instance segmentation, it does not possess the ability to track objects across multiple video frames.

Although MaskCut can effectively locate and segment multiple objects in an image, it operates only on a single image. When applied naively to a sequence, it lacks temporal consistency in the instance segmentation masks produced across video frames. 

\subsection{ImageCut2Video Synthesis for Training}
\label{sec:imgcut2vid}

We propose a learning-based approach to ensuring temporal consistency in video segmentation masks, based on generating synthetic videos from pairs of individual images and MaskCut masks. Surprisingly, we found that an extremely simple synthetic video generation method yields sufficient training data to learn a powerful video segmentation model that can operate on videos with much greater complexity of motion than is present in the training data.

Given unlabeled images in the minibatch and their pseudo-masks, our \vidsy method synthesizes corresponding videos and pseudo-mask trajectories, thereby allowing us to train the model in an unsupervised manner while offering the necessary supervision for simultaneous detection, segmentation, and tracking of objects in videos.

% Unlike prior unsupervised video segmentation methods that depend on pre-trained optical flow models and natural videos, our proposed approach can be trained exclusively on unlabeled images, such as \imnet.

First, given an image and its corresponding pseudo-masks in the mini-batch, we duplicate the image $t$ times and connect its MaskCut pseudo-masks to form the initial trajectories. This synthetic video, however, only contains static foreground objects.
To generate additional trajectories with mobile objects, a second image is randomly selected from the mini-batch, and its objects are cropped using its MaskCut pseudo-masks. These objects are then randomly resized, repositioned, and augmented before being pasted onto the first image. The resulting masks are connected along the temporal dimension to generate additional trajectories with mobile objects.

Specifically, given a target image $I_1$, a random source image $I_2$ in the mini-batch and its corresponding set of binary pseudo-masks $\{M_2^1,...,M_2^s\}$, we first apply a transformation function $\mathcal{T}$ to resize and shift these pseudo-masks randomly. 
This gives us a new set of pseudo-masks $\{\hat{M}_2^1,...,\hat{M}_2^s\}$, where $\hat{M}_2^s=\mathcal{T}({M}_2^s)$. 
Next, we synthesize a video with $t$ frames by duplicating image $I_1$ for $t$ times and pasting the augmented masks onto $I_1$ using:
\begin{align}
    I_1^t\!=\!I_1\times\Pi_{i=1}^{s}(1\!-\!\hat{M}_2^i)\!+\!I_2\times (1\!-\!\Pi_{i=1}^{s} (1-\hat{M}_2^i))
\end{align}
\noindent where $\times$ refers to element-wise multiplication. 
% We apply various resize scales and shift functions $\mathcal{T}$ to preprocess the pseudo masks for each frame in the synthetic video.
% By connecting the same pseudo mask across consecutive frames, we obtain a pseudo trajectory of the object in the synthetic video.
% Subsequently, these pseudo trajectories of objects can be used to train a video instance segmentation model in an unsupervised manner. 

\subsection{Video Segmentation Model}
\label{sec:model_learning}
During training, the resulting synthetic videos produced by \vidsy, comprising both mobile and stationary objects, are employed as the inputs to train a video instance segmentation model. The segmentation mask trajectories corresponding to each object in the video serve as `ground-truth' annotations.

% \tjd{add a paragraph describing the video segmentation model at a high level here}

We utilize VideoMask2Former~\cite{cheng2022masked,cheng2021mask2former} with a backbone of ResNet50~\cite{he2016deep} as our video instance segmentation (VIS) model.
It operates by attending to the 3D spatiotemporal features of our synthetic videos and generating 3D volume predictions of pseudo-mask trajectories using shared queries across frames. 
The shared queries across frames enable the model to segment and track object instances based on their appearance (feature) similarities, making it a powerful framework for analyzing video sequences. 

\section{Implementation Details}
\label{sec:implementation}

\noindent \textbf{\ours.} 
We first employ the MaskCut approach on images preprocessed to a resolution of 480$\times$480 pixels. 
We then compute a patch-wise cosine similarity matrix using the pretrained ViT-Base/8 DINO~\cite{caron2021emerging} model, which serves as input to the MaskCut algorithm for initial segmentation mask generation.
We set $t=3$, which is the maximum number of masks per image.
To refine the segmentation masks, we employ a post-processing step using Conditional Random Fields (CRFs)~\cite{krahenbuhl2011efficient}, which enforces smoothness constraints and preserves object boundaries, resulting in improved segmentation masks.

Next, we use \vidsy to synthetic videos given images and their pseudo-masks in a mini-batch.
We found that synthetic videos with two frames are sufficient to train a video instance segmentation model; therefore, we use $s\!=\!2$ by default.
We randomly change the brightness, contrast, and rotation of the masks to create new variations of pseudo-masks.
Additionally, we randomly resize the pseudo-masks (scale$\in$[0.8,1.0]), and shift their positions.
% For the DropLoss, we use $\tau\!=\!0.01$.

\tabUnsupYTVIS{t!}

\noindent \textbf{Training and test data.}
\label{sec:train-test-data}
Our model is trained solely on the unlabeled images from ImageNet~\cite{deng2009imagenet}, which comprises approximately 1.3 million images. 
Without further fine-tuning on any video datasets, we test our model's zero-shot unsupervised video instance segmentation performance on four multi-instance video segmentation benchmarks, including YouTubeVIS-2019~\cite{yang2019video}, YouTubeVIS-2021~\cite{yang2019video}, DAVIS2017~\cite{perazzi2016benchmark}, and DAVIS2017-Motion~\cite{perazzi2016benchmark,pont20172017}. 

YoutubeVIS-2019 and YouTube-VIS2021 contain 2,883 high-resolution YouTube videos (2,238 training videos and 302 validation videos) and 3,859 high-resolution YouTube videos (2,985 training videos and 421 validation videos), respectively. 
We evaluate the zero-shot unsupervised learning performance on their training splits in a class-agnostic manner. 
For DAVIS-2017, we evaluate our model's performance on the 30 videos from its validation set.
% It is worth noting that 12 out of the 30 videos from DAVIS2017 and 26 out of the 30 videos from DAVIS2017-Motion contain only one moving instance. Additionally, the DAVIS datasets concentrate exclusively on the performance of prominent moving objects, even in videos containing multiple objects, whether moving or static.

\noindent \textbf{Training settings.}
\noindent \textbf{1) Unsupervised Image Model Pretraining:} 
We first pretrain a Mask2Former~\cite{cheng2022masked} model with a backbone of ResNet50~\cite{he2016deep} on \imnet using MaskCut's pseudo-masks. 
The model is optimized for 160k iterations, with a batch size of 16 and a learning rate of 0.00002. 
The learning rate is decayed by a factor of 20 at iteration 80,000. 
To prevent overfitting, a dropout layer with a rate of 0.3 is added after the self-attention layers of transformer decoders.
\noindent \textbf{2) Unsupervised Video Model Learning:} 
We initialize the VideoMask2Former model~\cite{cheng2021mask2former} with model weights from the previous stage, and then fine-tune it on the synthetic videos we construct from \imnet.
We train \ours on 8 A100 GPUs for 80k iterations, using the AdamW optimizer~\cite{loshchilov2017decoupled}. We set the initial learning rate to 0.000005 and apply a learning rate multiplier of 0.1 to the backbone. A dropout layer with a rate of 0.3 is added after the self-attention layers of transformer decoders. 

\noindent \textbf{Evaluation metric AP$^{\text{video}}$ and AR$^{\text{video}}$}:
The evaluation metrics used in YouTubeVIS are Averaged Precision (AP) and Averaged Recall (AR), which are similar to those used in COCO~\cite{lin2014microsoft}. The evaluation is specifically conducted at 10 intersection-over-union (IoU) thresholds ranging from 50\% to 95\% with a step of 5\%~\cite{yang2019video}.
However, unlike in image instance segmentation, each instance in a video comprises a sequence of masks, so the IoU computation is performed not only in the spatial domain, but also in the temporal domain by summing the intersections at every single frame over the unions at every single frame.

\noindent \textbf{Evaluation metric $\mathcal{J}$ and $\mathcal{F}$}:
For DAVIS~\cite{pont20172017}, we report results using their official evaluation metrics $\mathcal{J}\&\mathcal{F}$, $\mathcal{J}$ and $\mathcal{F}$. 
The region measure ($\mathcal{J}$)~\cite{pont20172017} is the intersection-over-union (IoU) score between the algorithm's mask and the ground-truth mask. 
The boundary measure ($\mathcal{F}$)~\cite{pont20172017} is the average precision of the boundary of the algorithm's mask. 
The evaluation metrics are computed separately for each instance, and then the results are averaged over all instances to get the final score. 
$\mathcal{J}\&\mathcal{F}$ is the mean of $\mathcal{J}$ and $\mathcal{F}$.
\def\figOOD#1{
    \captionsetup[sub]{font=small}
    \begin{figure}[#1]
      \centering
      \includegraphics[width=1.0\linewidth]{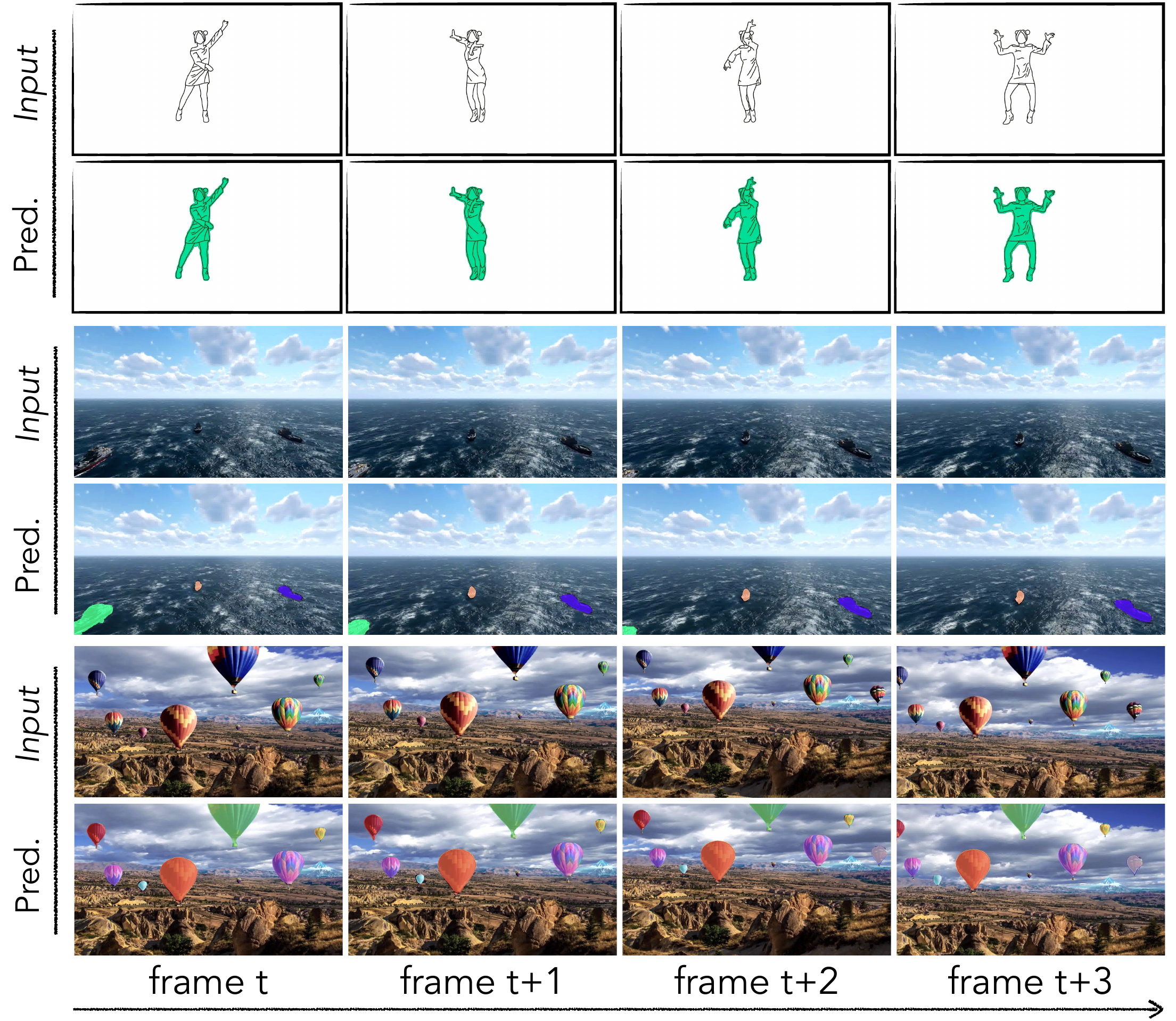}
      % \vspace{-6pt}
      \caption{
      We present qualitative results on videos covering a range of \textbf{out-of-domain} sources, including sketches, 3D computer-generated imagery (CGI), hybrid (CGI + realistic), \etc. 
      Our approach, \ours, can produce high-quality segmentation and tracking results for small objects that are often difficult to distinguish from the background, as well as for object sketches that lack textual information.
      % Input and Pred. refer to RGB video frames and model predictions, respectively.
      }
      \label{fig:ood}
    \end{figure}
}

\def\figSemiSL#1{
    \captionsetup[sub]{font=small}
    \begin{figure}[#1]
      \centering
      \includegraphics[width=1.0\linewidth]{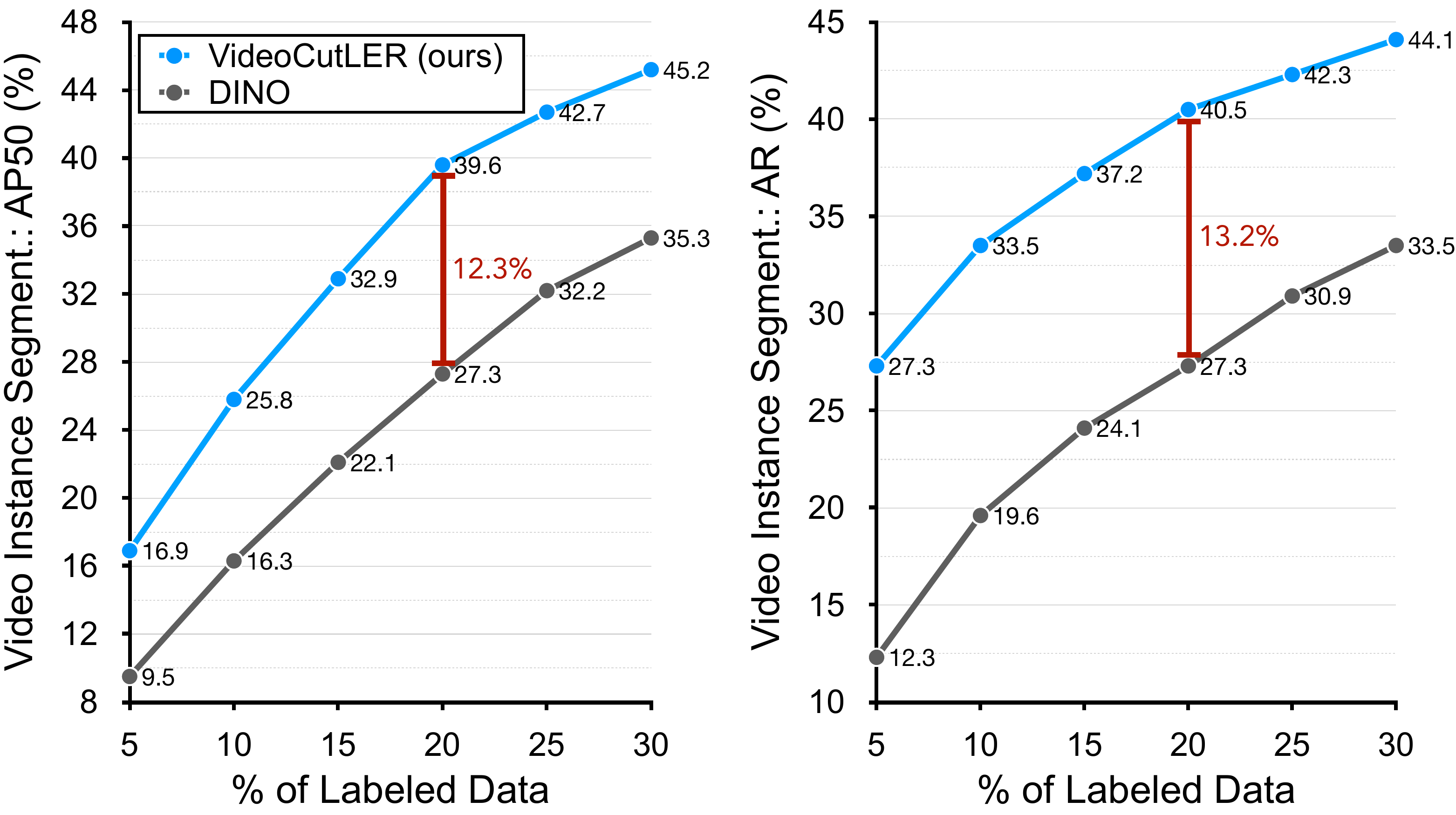}
      % \vspace{-8pt}
      \caption{
        We fine-tune \ours for \textbf{semi-supervised video instance segmentation} on the YouTubeVIS-2019 dataset, using different percentages of labeled training data. We evaluate the performance of our method by reporting the average precision and recall on the validation set of YouTubeVIS-2019. To establish a strong baseline, we use the self-supervised DINO~\cite{caron2021emerging} model and initialize the weights of VideoMask2Former with DINO. To ensure a fair comparison, both baselines and \ours are trained using the same schedule and recipe.
      }
      \label{fig:semi-sup}
    \end{figure}
}

%%%%%%%%%%%%%%%%%%%%%%%%%%%%%%%%%%%%%%%%%%%%%%%%%%%%%%%%%%%%%%%%%%%%%%%%
\def\tabSupYTVIS#1{
\begin{table*}[#1]
% \vspace{-16pt}
\tablestyle{3.8pt}{1.0}
\small
\begin{center}
\begin{tabular}{p{2.cm}p{2.3cm}x{0.9cm}x{0.9cm}x{0.9cm}lcccccclcccccc}
\Xhline{0.8pt}
\multirow{2}{*}{Methods} & \multirow{2}{*}{Architecture} & \multicolumn{6}{c}{{YouTubeVIS-2019}} && \multicolumn{6}{c}{{YouTubeVIS-2021}} \\
\cline{3-8} \cline{10-15}
&& \multicolumn{1}{c}{AP} & \multicolumn{1}{c}{AP$_{50}$} & \multicolumn{1}{c}{AP$_{75}$} & \multicolumn{1}{c}{AP$_S$} & \multicolumn{1}{c}{AP$_M$} & \multicolumn{1}{c}{AP$_L$} && \multicolumn{1}{c}{AP} & \multicolumn{1}{c}{AP$_{50}$} & \multicolumn{1}{c}{AP$_{75}$} & \multicolumn{1}{c}{AP$_S$} & \multicolumn{1}{c}{AP$_M$} & \multicolumn{1}{c}{AP$_L$} \\ [.1em]
\hline
DINO \cite{caron2021emerging} & Mask2Former~\cite{cheng2021mask2former} & \phantom{1}23.0 & \phantom{1}39.0 &  \phantom{1}23.7 & \phantom{1}\phantom{1}6.0 &  \phantom{1}28.0 &  \phantom{1}34.2 && 24.6 & \phantom{1}41.4 & \phantom{1}25.9 & \phantom{1}8.7 & 34.0 & \phantom{1}39.9 \\
\rowcolor{Gray}
\ours  & Mask2Former~\cite{cheng2021mask2former} &  \phantom{1}38.9 & \phantom{1}56.7 & \phantom{1}43.3 & \phantom{1}22.1 & \phantom{1}43.1 & \phantom{1}51.8 && 33.4 & \phantom{1}53.8 & \phantom{1}36.3 & 15.7 & 40.9 & \phantom{1}54.8 \\
\rowcolor{Gray}
\textit{vs. prev. SOTA} && \plus{+15.9} & \plus{+17.7} & \plus{+19.6} & \plus{+16.1} & \plus{+15.1} & \plus{+17.6} && \plus{+8.8}   & \plus{+12.4} & \plus{+10.4} & \plus{+7.0} & \plus{+6.9} & \plus{+14.9}  \\
\Xhline{0.8pt}
\end{tabular}
\end{center}
\vspace{-10pt}
\caption{VideoCutLER can serve as a strong pretrained model for the \textbf{supervised video instance segmentation} task. The video segmentation model, Mask2Former, is initialized with various pretrained models, \ie, DINO or \ours, and fine-tuned on the training set with human annotations. We report the instance segmentation metrics and evaluate the model performance on the \texttt{val} splits.
}
\label{tab:sup-ytvis}
\end{table*}
}

\def\tabVSSupYTVIS#1{
\begin{table*}[#1]
\tablestyle{4.2pt}{1.0}
\small
\begin{center}
\begin{tabular}{lx{0.9cm}x{0.9cm}x{0.8cm}x{2.6cm}lccccccc}
\Xhline{0.8pt}
\multirow{2}{*}{Methods} & \multicolumn{4}{c}{{Training settings}} && \multicolumn{7}{c}{{YouTubeVIS-2021$\setminus$YouTubeVIS-2019}}\\
\cline{2-5} \cline{7-13}
& flow & videos & sup. & training data && \multicolumn{1}{c}{AP$_{50}$} & \multicolumn{1}{c}{AP$_{75}$} & \multicolumn{1}{c}{AP} & \multicolumn{1}{c}{AP$_S$} & \multicolumn{1}{c}{AP$_M$} & \multicolumn{1}{c}{AP$_L$} & \multicolumn{1}{c}{AR$_{100}$} \\ [.1em]
\hline
\tc{MaskTrack R-CNN$^*$~\cite{yang2019video}} & \tc{\cmark} & \tc{\cmark} & \tc{\cmark} & \tc{\scriptsize IN-1K+YT2019} && \tc{\phantom{1}32.4} & \tc{\phantom{1}13.0} &  \tc{\phantom{1}15.0} & \phantom{1}\tc{8.4} &  \tc{\phantom{1}24.9} & \tc{\phantom{1}39.0} & \tc{\phantom{1}20.3} \\
\tc{MaskTrack R-CNN$^*$~\cite{yang2019video}} & \tc{\cmark} & \tc{\cmark} & \tc{\cmark} & \tc{\scriptsize IN-1K+COCO+YT2019} && \tc{\phantom{1}35.8} & \tc{\phantom{1}18.7} &  \tc{\phantom{1}18.7} & \tc{10.5} &  \tc{\phantom{1}31.3} & \tc{\phantom{1}46.8} & \tc{\phantom{1}24.5} \\
% \hline
OCLR$^*$~\cite{xie2022segmenting} & \cmark & \cmark & \xmark & IN-1K+synthetic && \phantom{1}\phantom{1}3.3 & \phantom{1}\phantom{1}0.2 & \phantom{1}\phantom{1}1.0 & \phantom{1}0.3 & \phantom{1}\phantom{1}2.7 & \phantom{1}\phantom{1}7.5 & \phantom{1}\phantom{1}5.4 \\
\rowcolor{Gray}
\ours & \xmark & \xmark & \xmark & IN-1K && \phantom{1}21.4 & \phantom{1}\phantom{1}7.1 & \phantom{1}\phantom{1}9.0 & \phantom{1}4.9 & \phantom{1}13.3 & \phantom{1}29.6 & \phantom{1}17.1  \\
\rowcolor{Gray}
\textit{vs. prev. SOTA} &&&&&& \plus{+18.1} & \phantom{1}\plus{+6.9} & \phantom{1}\plus{+8.0} & \plus{+4.6} & \plus{+10.6} & \plus{+22.1} & \plus{+11.7} \\
\Xhline{0.8pt}
\end{tabular}
\end{center}
\vspace{-10pt}
\caption{\ours greatly narrows the \textbf{gap between fully-supervised learning and unsupervised learning} for {multi-instance video segmentation}. Results are evaluated in a class-agnostic manner on the relative complement of the set of videos from YouTubeVIS-2021 and the set of videos from YouTubeVIS-2019.
$^*$: reproduced results with the official code and checkpoints. IN-1K refers to ImageNet-1K.
}
% \vspace{-16pt}
\label{tab:vs-sup-ytvis}
\end{table*}
}

\def\tabDavis#1{
\begin{table*}[#1]
\tablestyle{1.9pt}{1.0}
\small
\begin{center}
\begin{tabular}{lx{0.9cm}x{0.9cm}x{0.8cm}x{2.6cm}lccccccc}
\Xhline{0.8pt}
\multirow{2}{*}{Methods} & \multicolumn{4}{c}{{Training settings}} && \multicolumn{3}{c}{{DAVIS2017}} &&\multicolumn{3}{c}{{DAVIS2017-Motion}}\\
\cline{2-5} \cline{7-9} \cline{11-13}
& flow & videos & sup. & training data && \multicolumn{1}{c}{$\mathcal{J}\&\mathcal{F}$} & \multicolumn{1}{c}{$\mathcal{J}$(Mean)} & \multicolumn{1}{c}{$\mathcal{F}$(Mean)} && \multicolumn{1}{c}{$\mathcal{J}\&\mathcal{F}$} & \multicolumn{1}{c}{$\mathcal{J}$(Mean)} & \multicolumn{1}{c}{$\mathcal{F}$(Mean)} \\ [.1em]
\hline % 55.1 54.5 55.7
MotionGroup (sup.)~\cite{yang2021self} & \cmark & \cmark & \xmark & IN-1K+synthetic && - & - & - && 39.5 & 44.9 & 34.2 \\
Mask R-CNN (w/ flow)$^{*}$~\cite{he2017mask,xie2022segmenting} & \cmark & \cmark & \xmark & IN-1K+synthetic && - & - & - && 50.3 & 50.4 & 50.2 \\
OCLR (w/ flow)$^{*}$~\cite{xie2022segmenting} & \cmark & \cmark & \xmark & IN-1K+synthetic && 39.6 & 38.2 & 41.1 && 55.1 & 54.5 & 55.7 \\
% \hline
\rowcolor{Gray}
\ours & \xmark & \xmark & \xmark & IN-1K && 43.6 & 41.7 & 45.5 && 57.3 & 57.4 & 57.2  \\
\rowcolor{Gray}
\textit{vs. prev. SOTA} &&&&&& \plus{+4.0}  & \plus{+3.5} & \plus{+4.4} && \plus{+2.2} & \plus{+2.9} & \plus{+1.5} \\
\Xhline{0.8pt}
\end{tabular}
\end{center}
\vspace{-10pt}
\caption{\textbf{Zero-shot unsupervised single/few-instance segmentation}. \ours also outperforms the previous state-of-the-arts on DAVIS2017 and DAVIS2017-Motion. \textit{Note: 12 out of 30 videos from DAVIS2017 and 26 out of 30 videos from DAVIS2017-Motion contain only 1 moving instance. Additionally, DAVIS datasets focus solely on the performance of moving prominent objects, even in videos where multiple objects are present.}
This disadvantages our model since it can segment both static and moving objects and has not been exposed to any downstream videos during training.
$^{*}$: utilize optical flow predictions from RAFT~\cite{teed2020raft}, which is pretrained on external videos.
% $^*$: results reproduced with official codes and checkpoints.
All methods are evaluated in a zero-shot manner, \ie no fine-tuning on target videos. 
}
% \vspace{-16pt}
\label{tab:davis}
\end{table*}
}

\def\tabAblationsVideoCutLER#1{
\begin{table}[#1]
    \centering
    \subfloat[
    \textbf{Frame size}.
    \label{tab:ablate_frame_size}
    ]{
        \centering
        \begin{minipage}{0.35\linewidth}{\begin{center}
                    \tablestyle{2pt}{1.1}
                    \begin{tabular}{ccac}
                        \Xhline{0.8pt}
                        Size $\rightarrow$ & 180 & 360 & 480 \\ [.1em]
                        \hline
                        % AP$^{\text{box}}_{50}$ & 17.9 & 19.5 & 20.8 & 21.1 \\
                        AP$^{\text{video}}_{50}$ & 49.9 & 50.7 & 50.4 \\
                        \Xhline{0.8pt}
                    \end{tabular}
        \end{center}}\end{minipage}
    }
    \subfloat[
    \textbf{\# frames}.
    \label{tab:ablate_nframes}
    ]{
        \begin{minipage}{0.6\linewidth}{\begin{center}
                    \tablestyle{1.5pt}{1.1}
                    \begin{tabular}{cccac}
                        \Xhline{0.8pt}
                        \# frames $\rightarrow$ & CutLER$^{\dagger}$~\cite{wang2023cut} & 2 & 3 & 4 \\ [.1em]
                        \hline
                        AP$^{\text{video}}_{50}$ & 37.5 & 49.8 & 50.7 & 50.4 \\
                        \Xhline{0.8pt}
                    \end{tabular}
        \end{center}}\end{minipage}
    }\vspace{5pt}
    \\
    \subfloat[
    \textbf{Data augmentations for \vidsy.}
    \label{tab:ablate_augs}
    ]{
        \begin{minipage}{0.98\linewidth}{\begin{center}
                    \tablestyle{2.8pt}{1.1}
                    \begin{tabular}{cccccca}
                        \Xhline{0.8pt}
                        Augmentations $\rightarrow$ & none & +bright &  +rotation & +contrast & +crop & all \\ [.1em]
                        \hline
                        AP$^{\text{video}}_{50}$ & 47.8 & 48.1 & 48.9 & 48.3 & 48.7 & 50.7 \\
                        \Xhline{0.8pt}
                    \end{tabular}
        \end{center}}\end{minipage}
    }
    % \\
    % \subfloat[
    % \textbf{Contribution of \vidsy}.
    % \label{tab:ablate_i2v}
    % ]{
    %     \begin{minipage}{0.98\linewidth}{\begin{center}
    %                 \tablestyle{2.0pt}{1.1}
    %                 \begin{tabular}{lcccccc}
    %                     \Xhline{0.8pt}
    %                      & AP$^{\text{video}}_{50}$ & AP$^{\text{video}}$ & AP$^{\text{video}}_{\text{S}}$ & AP$^{\text{video}}_{\text{M}}$ & AP$^{\text{video}}_{\text{L}}$ & AR$^{\text{video}}_{100}$ \\
    %                     \hline
    %                     CutLER$^{\dagger}$~\cite{wang2023cut} & \phantom{1}37.5 & \phantom{1}17.1 & \phantom{1}3.3 & \phantom{1}13.9 & \phantom{1}27.6 & \phantom{1}30.4 \\
    %                     \rowcolor{Gray}
    %                     +\vidsy & \phantom{1}50.3 & \phantom{1}25.1 & \phantom{1}5.3 & \phantom{1}20.6 & \phantom{1}36.7 & \phantom{1}41.2 \\ [.1em]
    %                     \Xhline{0.8pt}
    %                 \end{tabular}
    %     \end{center}}\end{minipage}
    % }
    \vspace{-4pt}
    \caption{\textbf{Ablations for \ours}. We report video instance segmentation result AP$^{\text{video}}_{50}$ on YoutubeVIS-2019.
    \textbf{(a)} We analyze the impact of varying the size of video frames on training \ours.
    \textbf{(b)} We evaluate the effect of the number of frames used for training video instance segmentation models.
    \textbf{(c)} We investigate the contribution of several augmentation methods, including brightness, rotation, contrast, and random cropping, which are used as default during model training.
    % \textbf{(d)} We demonstrate the effectiveness of \vidsy in training unsupervised multi-instance video segmentation models.
    % $^{\dagger}$: We train a CutLER~\cite{wang2023cut} model with Mask2Former as a detector on \imnet, which is regarded as single-frame videos, following its official training recipe.
    Default settings are highlighted in \colorrowtext{}.}
    \label{tab:ablate}
\end{table}
}

\def\figControlNet#1{
    \captionsetup[sub]{font=small}
    \begin{figure}[#1]
      \centering
      \includegraphics[width=1.0\linewidth]{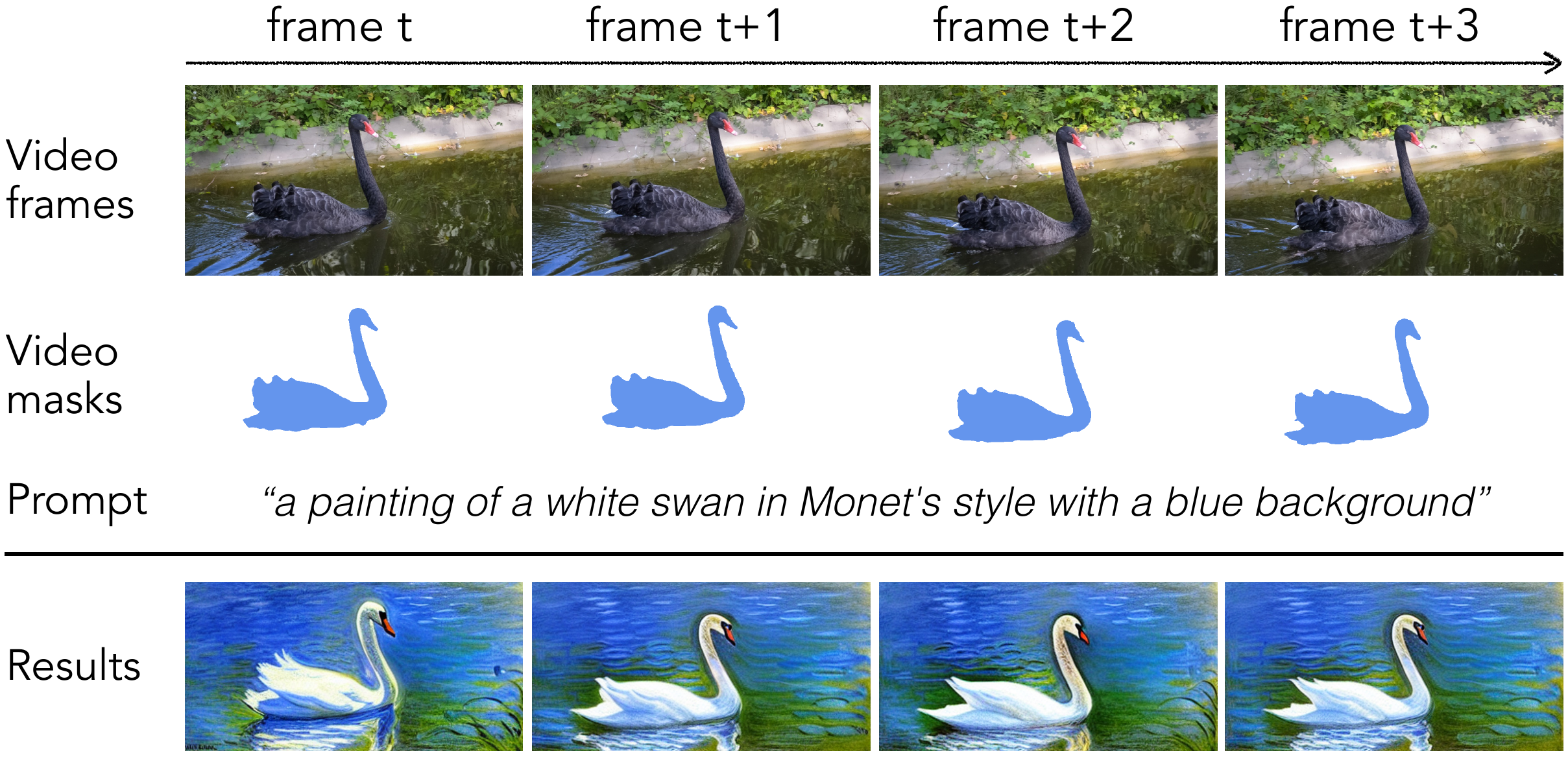}\vspace{-4pt}
      \caption{
        Controlling Stable Diffusion~\cite{zhang2023adding} using video instance segmentation masks produced by \ours. 
        This enables instance-consistent video editing across video frames, with the original instance shapes preserved.
      }
      \label{fig:controlnet}
    \end{figure}
}

\def\tabExtraVis#1{
\captionsetup[sub]{font=small}
\begin{figure}[#1]
  \centering
  \includegraphics[width=1.0\linewidth]{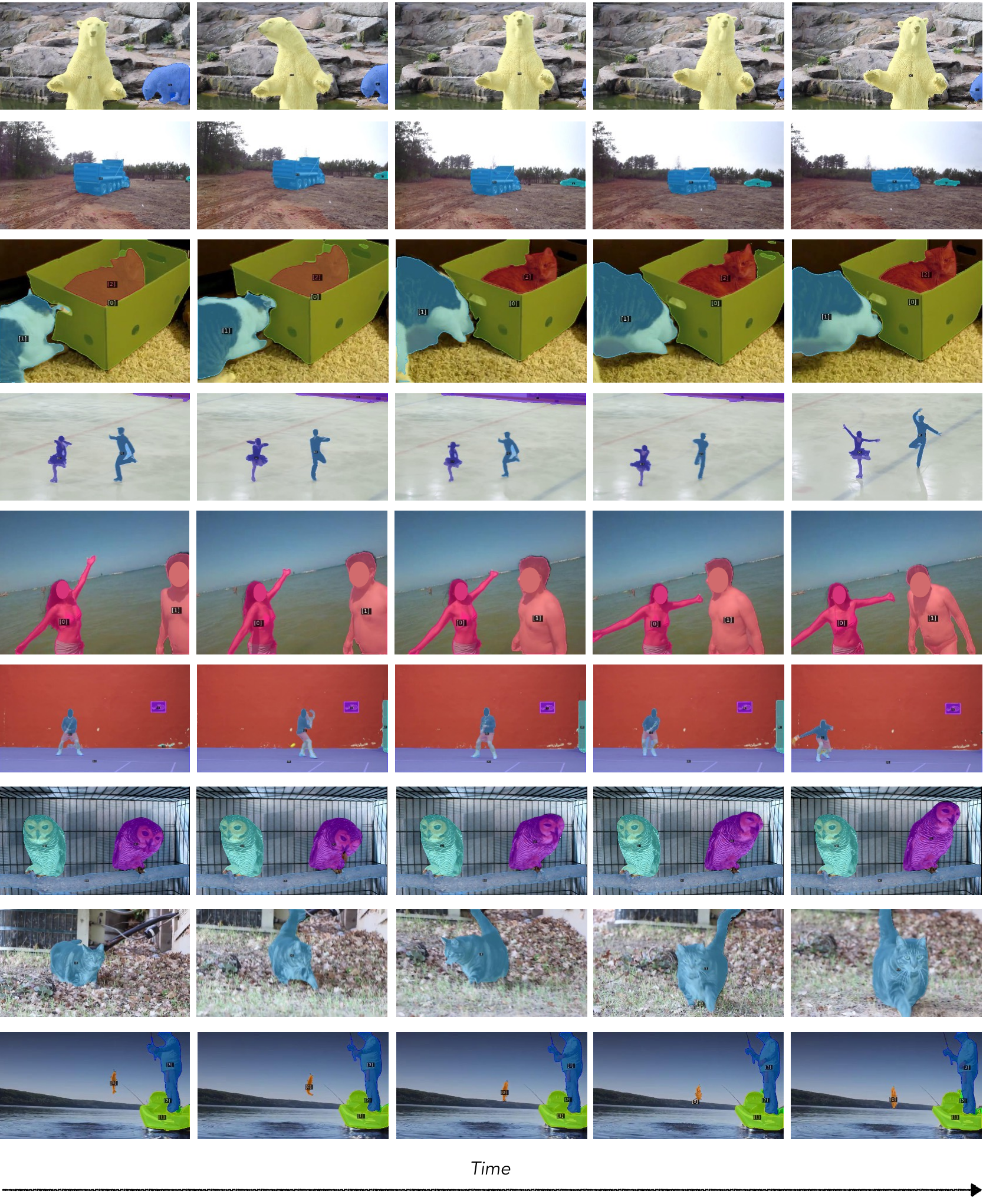}
  \caption{
   We present \textbf{qualitative visualizations} illustrating the zero-shot unsupervised video instance segmentation outcomes of VideoCutLER on YouTubeVIS dataset. 
   It's noteworthy that VideoCutLER is solely pretrained on image dataset ImageNet-1K, and its evaluation is conducted directly on the video dataset YouTubeVIS (no further fine-tuning required).
   The visual results provided effectively highlight that VideoCutLER is capable of segmenting and tracking multiple instances, delivering consistent tracking results across video frames, and successfully distinguishing between various instances, even when significant overlapping occurs.
   We show more demo results in appendix.
  }
  \label{fig:demo-ytvis}
\end{figure}
}

%%%%%%%%%%%%%%%%%%%%%%%%%%%%%%%%%%%%%%%%%%%%%%%%%%%%%%%%%%%%%%%%%%%%%%%%
\tabDavis{t!}

\section{Experiments}
\label{experiments}
We evaluate the performance of \ours on several video instance segmentation benchmarks. In~\cref{sec:exp-unuspervised-zs-seg}, we demonstrate that our approach can effectively perform segmentation and tracking of multiple objects in videos, even when trained on unlabeled \imnet images without any form of supervision. Our experimental results reveal that our method can drastically reduce the performance gap between unsupervised and supervised learning methods for video instance discovery and tracking. 
Furthermore, \cref{sec:exp-finetune-eval} demonstrates that fine-tuning \ours leads to further performance gains in video instance segmentation, surpassing previous works such as DINO in both fully supervised learning and semi-supervised learning tasks. In \cref{sec:ablation-study}, we conduct an ablation study to examine the impact of key components and their hyperparameters on the performance of our approach.

%%%%%%%%%%%%%%%%%%%%%%%%%%%%%%%%%%%%%%%%%%%%%%%%%%%%%%%%%%%%%%%%%%%%%%%%

\subsection{Unsupervised Zero-shot Evaluations}
\label{sec:exp-unuspervised-zs-seg}
In this section, we evaluate the performance of our method against previous state-of-the-art approaches on various video instance segmentation benchmarks. 

\par \noindent \textbf{Evaluating unsupervised video instance segmentation} poses two main challenges. 
Firstly, as unsupervised learning methods train the model without semantic classes, the class-aware video segmentation setup cannot be used directly for an evaluation. As a result, following previous works, we evaluate video instance segmentation results in a class-agnostic manner. 
Secondly, video instance segmentation datasets often annotate only a subset of the objects in the video, which makes Average Recall (AR) a valuable metric that does not penalize models for detecting novel objects not labeled in the dataset~\cite{wang2023cut}. 
Therefore, we report both AR and AP for YouTubeVIS.
Regarding DAVIS, we use the official unsupervised learning metrics $\mathcal{J}$, $\mathcal{F}$, and $\mathcal{J}\&\mathcal{F}$. 
All these metrics assess the performance of unsupervised video instance segmentation in a class-agnostic manner. 
\cref{sec:implementation} lists more details on evaluation metrics.

\par \noindent \textbf{Detailed comparisons on YouTubeVIS}.
\cref{tab:zero-shot-ytvis} presents a summary of the results for unsupervised zero-shot video instance segmentation on the YouTubeVIS-2019 and YouTubeVIS-2021 datasets. We compare our method's results with the previous state-of-the-art methods OCLR~\cite{xie2022segmenting} and motion grouping~\cite{yang2021self}. We reproduce their results using their official code and checkpoints to ensure fairness.

Although OCLR~\cite{xie2022segmenting} is also trained on synthetic videos, it relies on the off-the-shelf optical flow estimator RAFT~\cite{teed2020raft} to compute optical flows for RGB sequences. It is worth noting that RAFT is pretrained on a combination of synthetic videos~\cite{dosovitskiy2015flownet,Butler:ECCV:2012} and human-annotated videos such as KITTI-2015~\cite{keuper2015motion} and HD1K~\cite{kondermann2016hci}. 
Our approach, \ours, despite not using any optical flow estimations like many previous works on unsupervised video segmentation, achieves over 10$\times$ higher AP$_{\text{50}}$ and 18$\times$ higher AP than OCLR~\cite{xie2022segmenting} on YouTubeVIS-2019. Additionally, we achieve over 30\% higher recall. 
Furthermore, unlike the previous state-of-the-art method OCLR~\cite{xie2022segmenting}, which exhibits poor performance in segmenting small objects (with 0.0\% AP$_S$), our approach significantly outperforms it.
Similar performance gains can be observed on YouTubeVIS-2021.
Finally, the performance gains to CutLER~\cite{wang2023cut} demonstrates the effectiveness of \ours in training unsupervised multi-instance video segmentation models, surpassing CutLER by over 12.8\% on YouTubeVIS-2019.

In ~\fig{demo-ytvis}, we present qualitative visualizations illustrating the zero-shot unsupervised video instance segmentation outcomes of VideoCutLER on YouTubeVIS dataset.

\tabExtraVis{bt!}

\par \noindent \textbf{Detailed comparisons on DAVIS.}
To provide a comprehensive evaluation and comparison with existing unsupervised video instance segmentation approaches, we also assess the performance of our model on the validation sets of DAVIS-2017 and DAVIS2017-Motion~\cite{pont20172017,xie2022segmenting}.
Note that both DAVIS2017 and DAVIS2017-Motion datasets focus only on the performance of instance segmentation on \textit{\textbf{prominent moving objects}}, even in videos with multiple objects. As a result, only a single or a few objects of interest per video are annotated, which may not reflect the challenges that arise when multiple objects are present.

Although the evaluation of DAVIS is an unfair assessment for us since \ours is supposed to segment both static and moving objects, whereas DAVIS focuses on moving prominent objects, with only a single or a few moving objects of interest per video annotated. However, \cref{tab:davis} shows that \ours yields approximinately 4\% higher $\mathcal{J}$, $\mathcal{F}$, and $\mathcal{J}\&\mathcal{F}$. 
The additional results on DAVIS demonstrate that \ours achieves superior performance not only on static or minimally moving objects but also on dynamic objects, where prior methods relying on optical flow estimates can benefit from additional cues.

\tabVSSupYTVIS{t}
\par \noindent \textbf{Comparison of supervised and unsupervised learning} in object discovery and tracking abilities is presented in \cref{tab:vs-sup-ytvis}. We train a supervised MaskTrack R-CNN~\cite{yang2019video} model on the human-annotated training set of YouTubeVIS-2019 dataset, and evaluate it in a class-agnostic manner on the videos that are not shared between YouTubeVIS-2019 and YouTubeVIS-2021 datasets~\cite{yang2019video}. \cref{tab:vs-sup-ytvis} shows that our \ours model significantly narrows the gap between supervised learning and unsupervised learning methods in terms of the averaged precision AP$_{50}$ (gaps: 29.1\%$\rightarrow$11.0\%) and the averaged recall AR$_{100}$ (gaps: 14.9\%$\rightarrow$3.2\%), particularly for the AR$_{100}$.

\tabSupYTVIS{t}

\figSemiSL{t!}

\subsection{Label-Efficient and Fully-Supervised Learning} 
\label{sec:exp-finetune-eval}
In this section, we investigate \ours as a pretraining approach for supervised video instance segmentation models, and evaluate its effectiveness in label-efficient and fully-supervised learning scenarios.

\par \noindent \textbf{Setup.}
We use VideoMask2Former with a backbone of ResNet50 for all experiments in this section unless otherwise noted. 
For our experiments on semi-supervised learning, we randomly sample a subset of videos from the training split with different proportions of labeled videos. After pretraining our \ours model on \imnet, we fine-tune the model on the YouTubeVIS-2019~\cite{yang2019video} dataset with its human annotations.
For our experiments on the fully-supervised learning task, we fine-tune the \ours model on all available labeled data from the training sets of YouTubeVIS.
For baselines, we initialize a VideoMask2Former model with a DINO~\cite{caron2021emerging} model pre-trained on \imnet and fine-tuned on labeled videos. Since DINO has shown strong performance in detection and segmentation tasks, it serves as a strong baseline for our experiments. 

For semi-supervised learning, both the baselines and our models are trained for $2\times$ schedule, with a learning rate of 0.0001 for all model weights, except for the final classification layers, which use a learning rate of 0.0016. 
We train the models using a batch size of 16 and 8 GPUs.
For fully-supervised learning, we use the $1\times$ schedule and a learning rate of 0.0002 for the final classification layers.
We evaluate their performance on the \texttt{val} split of the YouTubeVIS-2019, and report results from its official evaluation server.

\noindent \textbf{Data for fully-/semi-supervised VIS.}
We fine-tune the pretrained VideoCutLER model on all or a subset of the training split of YouTubeVIS-2019. 
We then evaluate the resulting models on the validation set. To ensure a fair comparison, we use the same amount of human annotations to train our model and baselines. Specifically, we initialize the baselines with the DINO-pretrained model and fine-tune them on the training set of the respective dataset.
We evaluate the model performance on their validation sets and report results from its official evaluation server. 

\par \noindent \textbf{Results.}
Most prior approaches on self-supervised representation learning~\cite{he2020momentum,chen2020simple,wang2021unsupervised,grill2020bootstrap,caron2021emerging} are limited to providing initializations only for the model backbones, with the remaining layers, such as Mask2Former's decoders, being randomly initialized.
In contrast, \ours takes a more comprehensive approach that allows all model weights to be pretrained, resulting in a stronger pretrained model better suited for supervised learning.
As a result, as shown in \cref{fig:semi-sup} and \cref{tab:sup-ytvis}, our method outperforms these prior works significantly, offering a strong pretrained model for fully-/semi-supervised learning tasks.

In \cref{fig:semi-sup}, it can be observed that \ours consistently outperforms the strong baseline method DINO~\cite{caron2021emerging} across all label-efficient learning settings with varying proportions of labeled YouTubeVIS-2019 videos. The most significant performance gains are observed when 20\% labeled data is provided, where \ours exceeds DINO by over 12\% in video instance segmentation precision AP$_{50}$ and 13.2\% in terms of video instance segmentation recall.

As demonstrated in \cref{tab:sup-ytvis}, training the model with all available labeled videos from YouTubeVIS yields considerable performance improvements, surpassing DINO in terms of AP by more than 15.9\% on YouTubeVIS-2019 and 8.8\% on YouTubeVIS-2021, respectively.

\figOOD{!t}
\subsection{Ablation Study}
\label{sec:ablation-study}
\tabAblationsVideoCutLER{t!}
\par \noindent \textbf{Hyper-parameters and design choices.}
We present an ablation study on several key hyper-parameters and design choices of \ours in \cref{tab:ablate}. First, we analyze the impact of varying the size of video frames used for training \ours. From \cref{tab:ablate_frame_size}, we observe that the shortest edge length of 240 pixels yields the best performance. Using a larger resolution does not always lead to better results.
Next, \cref{tab:ablate_nframes} shows the effect of the number of frames used for training video instance segmentation models. We found that synthetic videos with three frames are optimal for learning an unsupervised video instance segmentation model. 
Increasing the number of frames does not result in a further improved performance, aligning with the findings reported in ~\cite{cheng2021mask2former}.
Furthermore, \cref{tab:ablate_augs} investigates the contribution of several augmentation methods, including brightness, rotation, contrast, and random cropping, which are used as default during model training. We found that compared to \vidsy without any data augmentations, adding these augmentations can bring about 3\% performance gains.

\par \noindent \textbf{Generalizability.} 
The qualitative results presented in \cref{fig:ood} demonstrate that \ours can effectively perform video instance segmentation and tracking on out-of-domain data sources, \eg sketches, 3D computer-generated imagery (CGI), and hybrid videos that combine CGI and real videos. These results showcase the generalizability of our model, which can be applied to a broad range of videos beyond the domains it was initially trained on, \ie, \imnet.

% \subsection{\ours for Video Editing}
% \label{sec:video-editing}
% \figControlNet{t!}
% In addition to the discriminative tasks presented in previous sections, in \cref{fig:controlnet}, we demonstrate that \ours can also be used for generative tasks, specifically video editing with Controlling Stable Diffusion using mask trajectories produced by \ours.
% Given a sequence of video frames, a sequence of masks produced by \ours, and a text prompt `\textit{a painting of a white swan in Monet's style with a blue background}', we can leverage a pretrained ControlNet~\cite{zhang2023adding} model for video editing that adheres to the specified conditions. With the aid of precise segmentation masks, the resulting output can accurately preserve the shape of the original instances in the video frames.
\section{Summary and Limitations}
We presented a simple unsupervised approach to segment multiple instances in a video.
Our approach, \ours, does not require labels, and does not rely on motion-based learning signals like optical flow.
In fact, \ours does not need real videos for training as we synthesize videos using natural images from the ImageNet-1K dataset.
Despite being simpler, \ours outperforms models that use additional learning signals or video data, achieving $10\times$ their performance on benchmarks like YouTubeVIS.
Additionally, \ours serves as a strong self-supervised pretrained model for supervised video instance segmentation.
We hope that our approach enables both a wide range of applications in video recognition, as well as its simplicity enables easy future research.

Limitations: while VideoCutLER demonstrates its capability to achieve the state-of-the-art performance without relying on optical flow estimations, potential further improvements may be obtained by leveraging natural videos and integrating joint training with optical flow estimations. However, for the sake of maintaining simplicity, we have chosen to defer these aspects to future research endeavors.

\textbf{Acknowledgement}
Trevor Darrell and XuDong Wang were funded by DoD including DARPA LwLL and the Berkeley AI Research Commons.

% Video segmentation is a fundamental task in computer vision, which enables a wide range of practical applications that require a deep understanding of the contents of video data.
% In this work, we tackle the challenging unsupervised multi-instance video segmentation task and present a frustrating simple cut-synthesize-and-learn pipeline, named \ours, that does not require any labels, neither pixel-level nor image-level annotations. 
% \ours is solely trained on unlabeled images, such as ImageNet-1K, yet it is capable of segmenting subsequent instances across video frames. Moreover, it outperforms existing approaches that rely on optical flow or motion estimation modules that have been pretrained on video data.
% Benchmarked on YouTubeVIS-2019 and YouTubeVIS-2021, VideoCutLER sets a new state-of-the-art (SOTA) performance with 47.8\% AP$_{50}$ and 36.8\% AP$_{50}$, respectively. This represents a significant improvement over the previous SOTA, surpassing it by 43\% (47.8\% vs. 4.8\%) and 32.4\% (36.8\% vs. 4.4\%), respectively. 
% In addition, we demonstrate that \ours can serve as a strong pretrained model for supervised video instance segmentation.

{
    \small
    \bibliographystyle{ieeenat_fullname}
    \bibliography{main}
}

% \clearpage
%%%%%%%%%%%%%%%%%%%%%%%%%%%%%%%%%%%%%%%%%%%%%%%%%%%%%%%%%%%%%%%%%%%%%%%%

% \def\tabExtraVis#1{
% \captionsetup[sub]{font=small}
% \begin{figure*}[#1]
%   \vspace{-15pt}
%   \centering
%   \includegraphics[width=0.95\linewidth]{figures/demos-more.pdf}
%   \caption{
%    In the appendix, we present additional qualitative visualizations (Figure A1-A6) illustrating the zero-shot unsupervised video instance segmentation outcomes of VideoCutLER on YouTubeVIS dataset. 
%    It's noteworthy that VideoCutLER is solely pretrained on image dataset ImageNet-1K, and its evaluation is conducted directly on the video dataset YouTubeVIS (no further fine-tuning required).
%    The visual results provided effectively highlight that VideoCutLER is capable of segmenting and tracking multiple instances, delivering consistent tracking results across video frames, and successfully distinguishing between various instances, even when significant overlapping occurs.
%   }
%   \label{fig:demo-a1}
% \end{figure*}
% }

\def\tabExtraVis#1{
    \centering \hspace{6pt} \vspace{15pt}
    \begin{minipage}{2\linewidth}% to keep image and title on one page
    \makebox[1\linewidth]{
        \centering
        \includegraphics[width=1\linewidth]{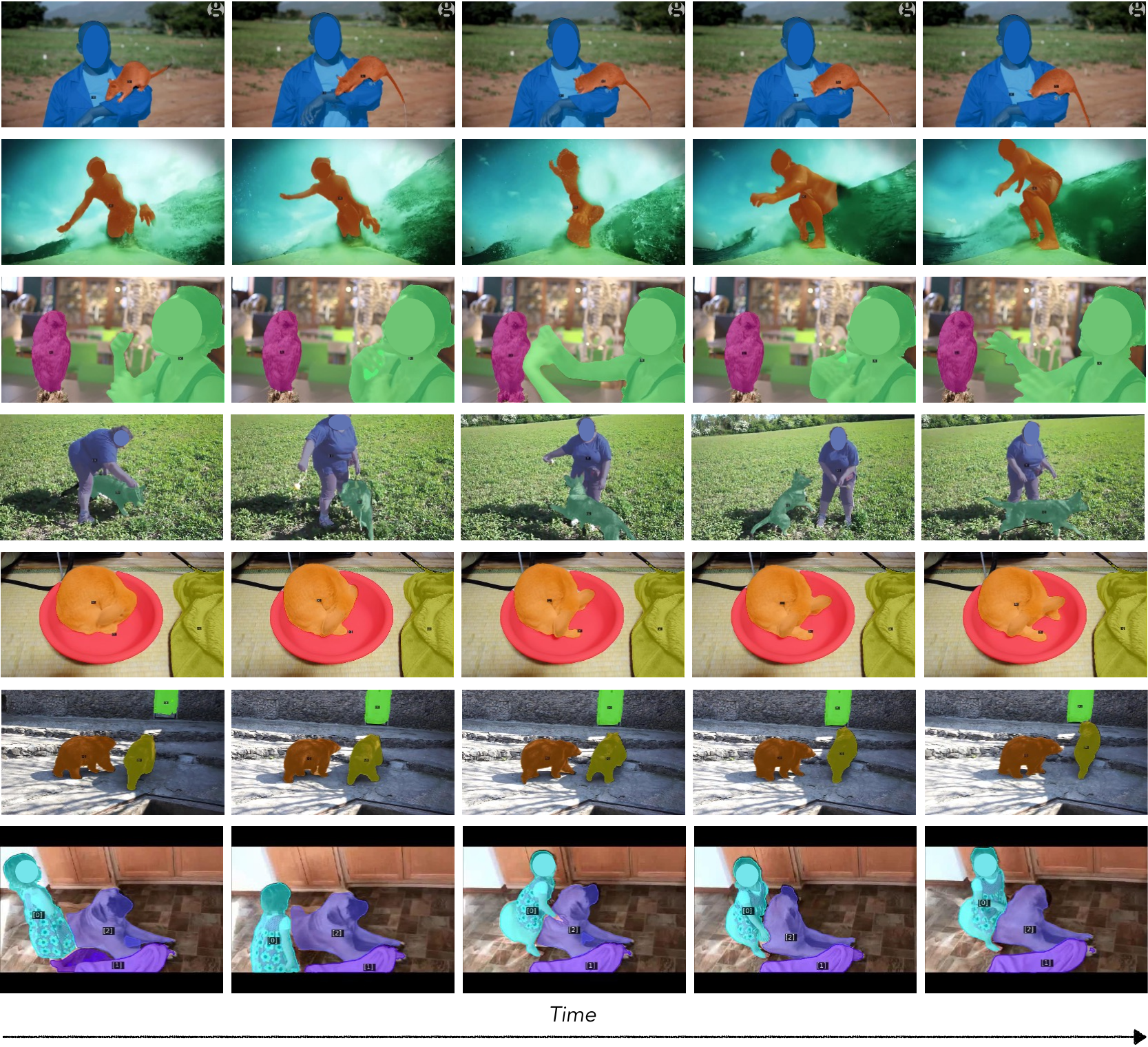}
    }
    \captionof{figure}{
    We present additional qualitative visualizations (Figures A1-A5) showcasing the zero-shot unsupervised video instance segmentation results of \ours on YoutubeVIS. 
    \ours is pretrained on image dataset \imnet-1K and directly evaluated on video dataset YouTubeVIS.
    }
    \label{fig:demo-a1}
    \end{minipage}
}

\def\tabExtraVisFive#1{
\captionsetup[sub]{font=small}
\begin{figure*}[#1]
  \vspace{15pt}
  \centering
  \includegraphics[width=0.95\linewidth]{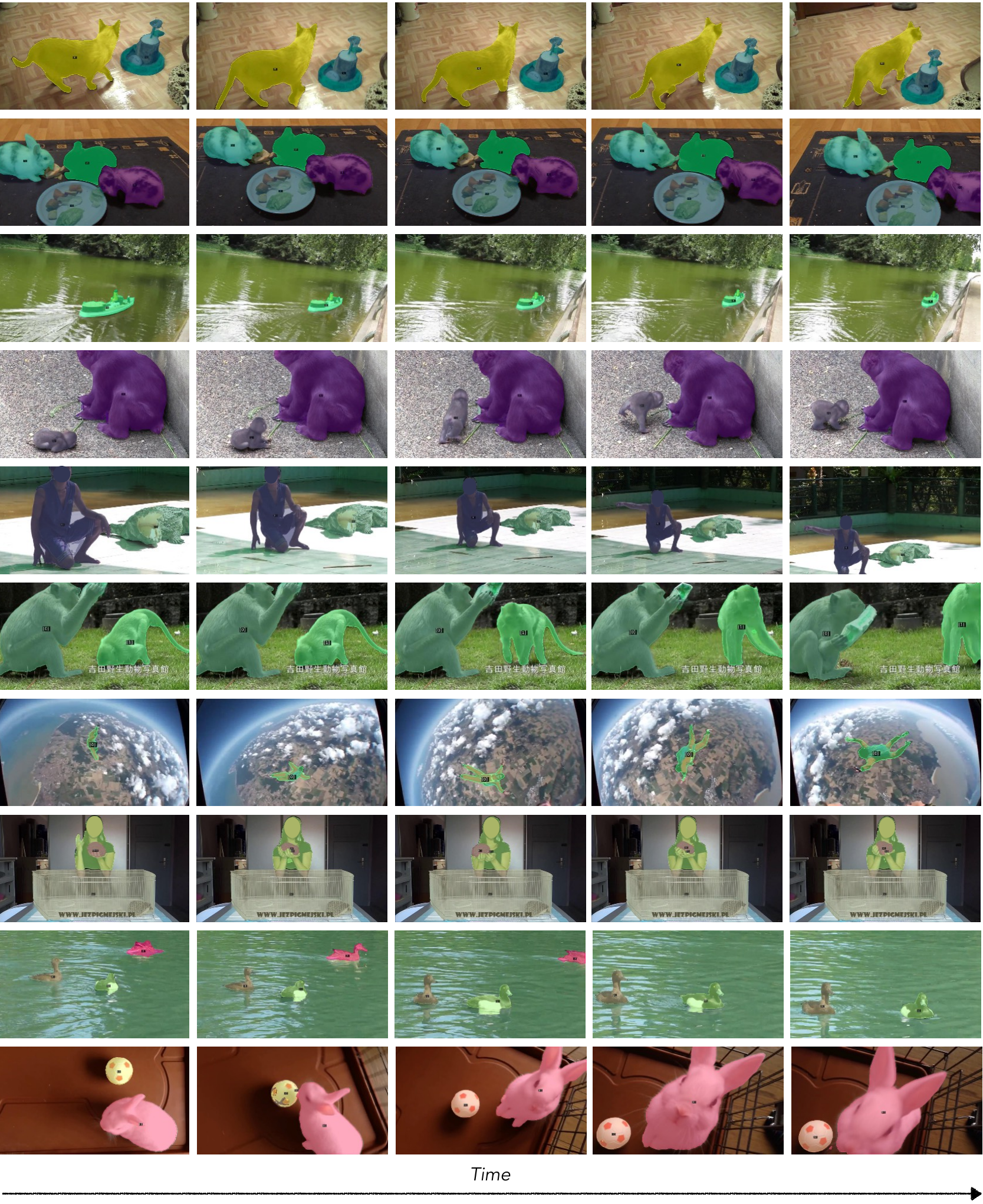}
  \caption{
    Additional qualitative visualizations.
  }
    \label{fig:demo-a2}
\end{figure*}
}

\def\tabExtraVisSix#1{
\captionsetup[sub]{font=small}
\begin{figure*}[#1]
  \vspace{15pt}
  \centering
  \includegraphics[width=0.95\linewidth]{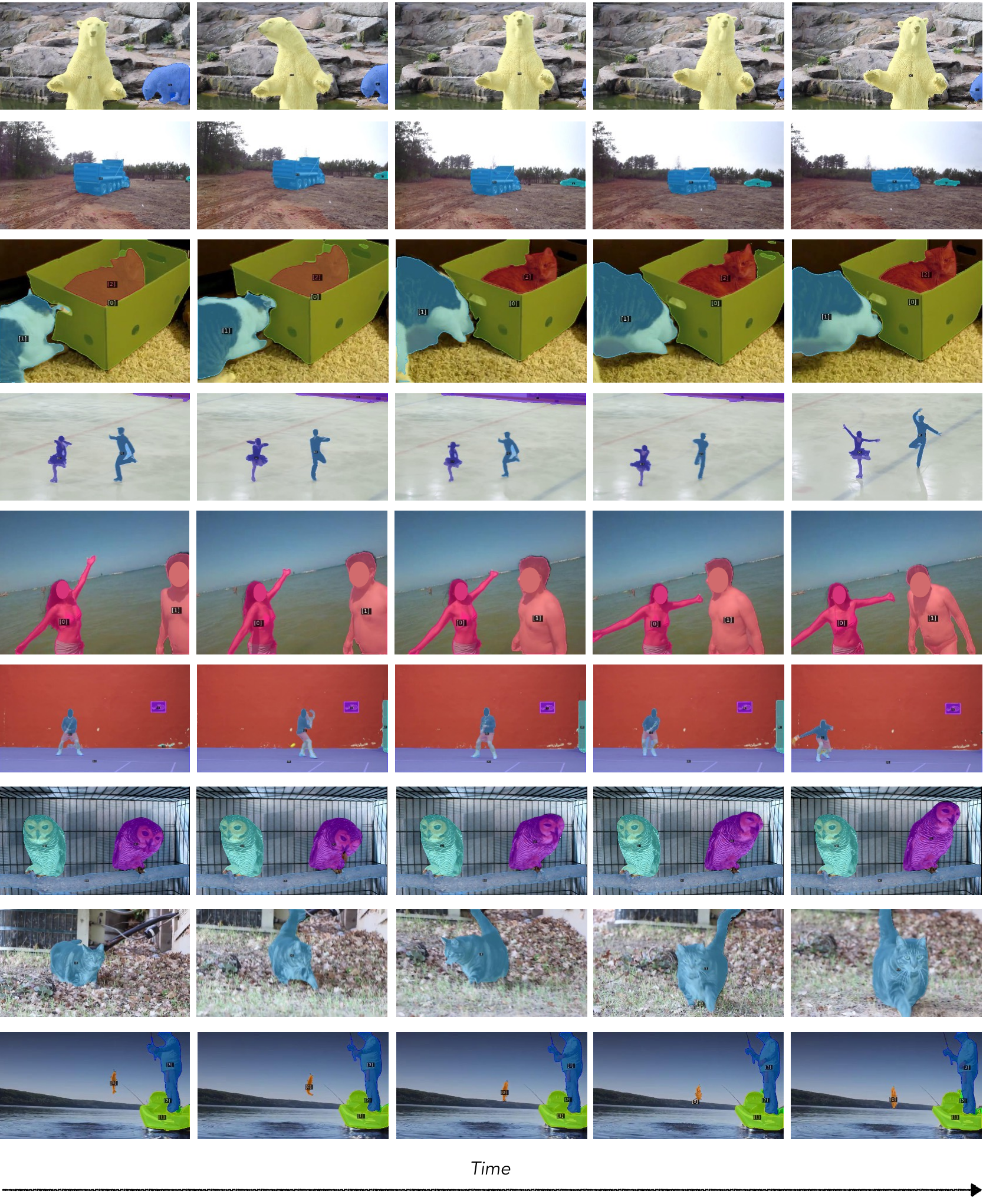}
  \caption{
    Additional qualitative visualizations.
  }
  \label{fig:demo-a4}
\end{figure*}
}

\def\tabExtraVisFour#1{
\captionsetup[sub]{font=small}
\begin{figure*}[#1]
  \vspace{15pt}
  \centering
  \includegraphics[width=0.95\linewidth]{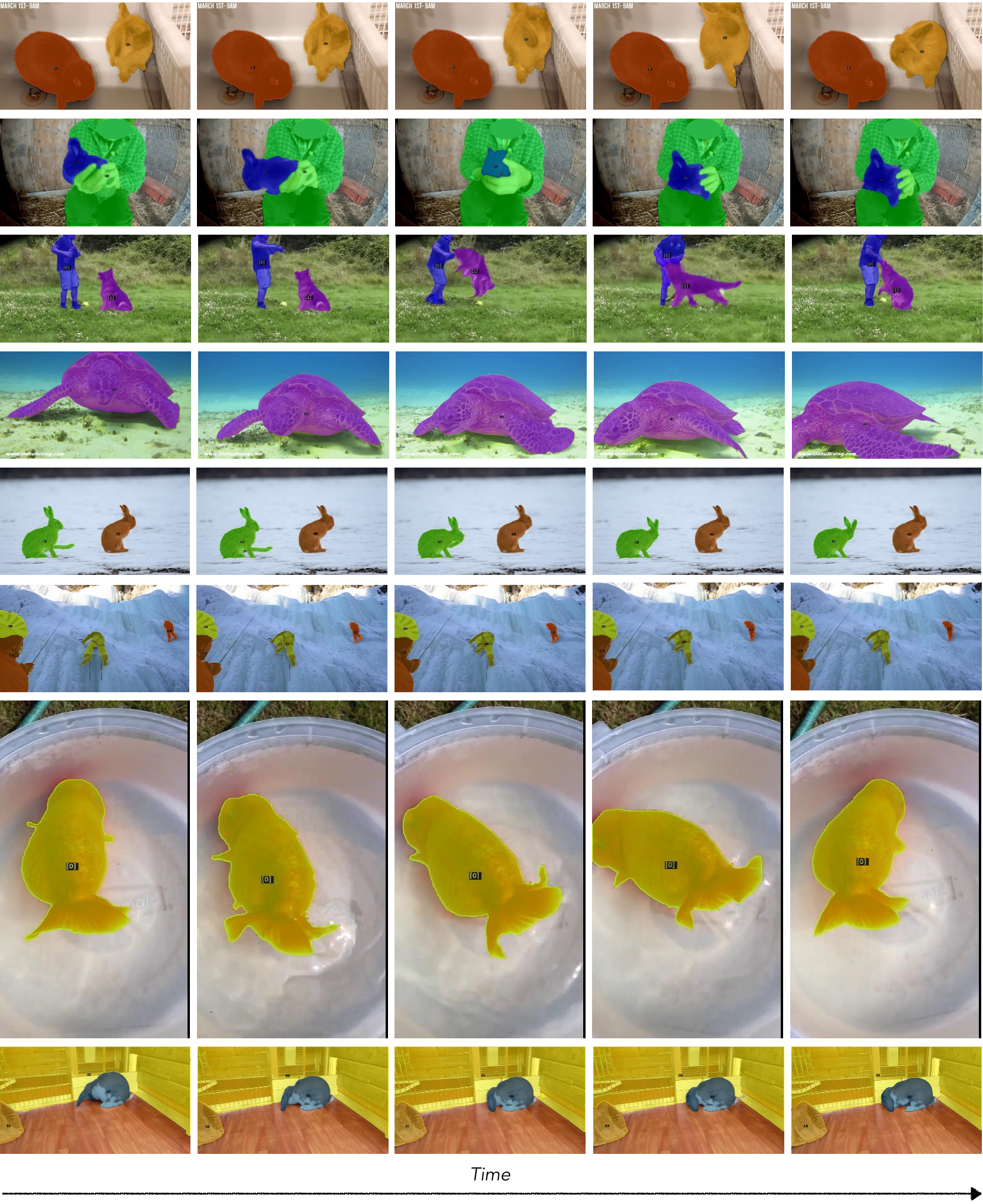}
  \caption{
    Additional qualitative visualizations.
  }
  \label{fig:demo-a3}
\end{figure*}
}

\def\tabExtraVisThree#1{
\captionsetup[sub]{font=small}
\begin{figure*}[#1]
  \vspace{15pt}
  \centering
  \includegraphics[width=0.95\linewidth]{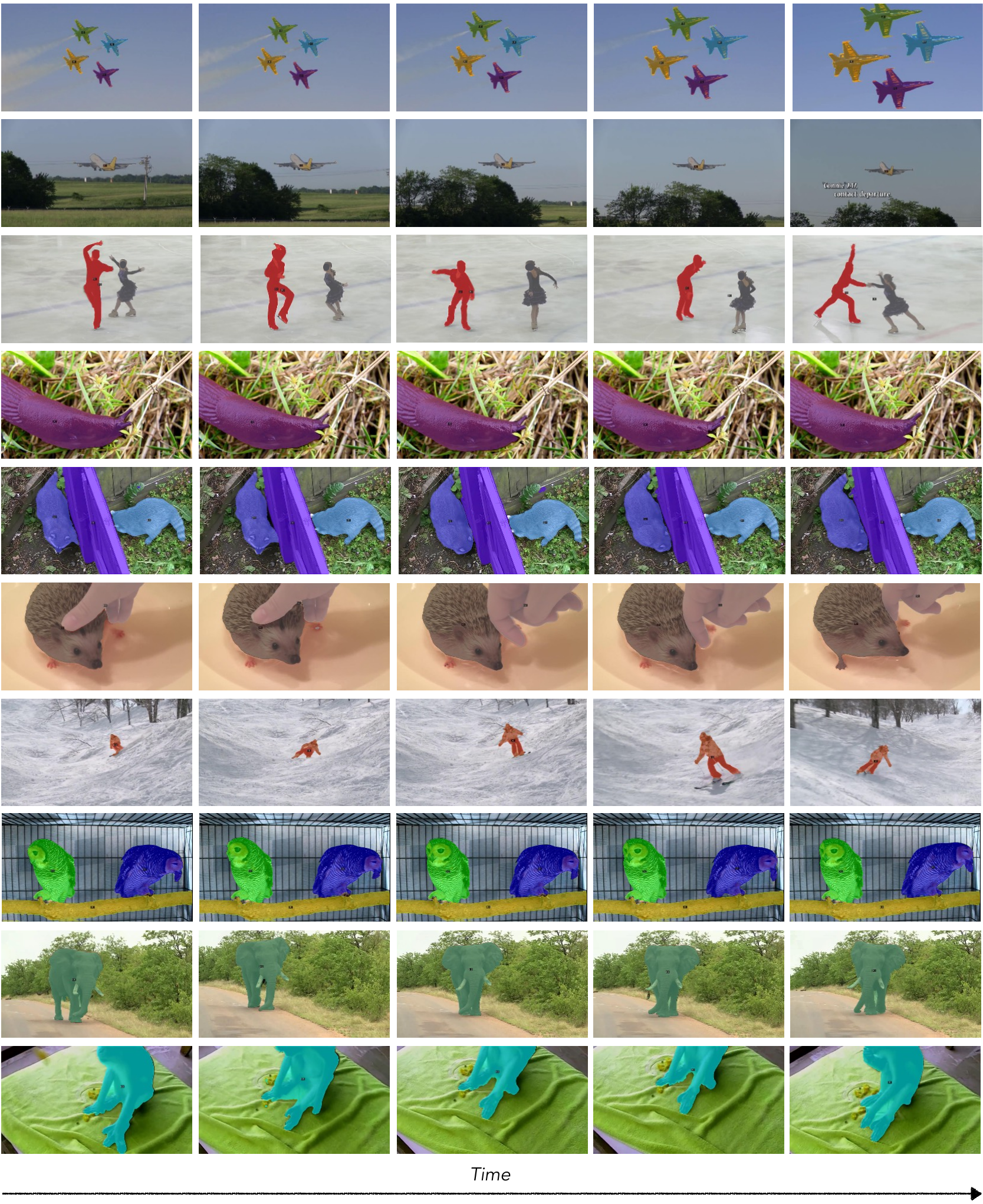}
  \caption{
    Additional qualitative visualizations.
  }
  \label{fig:demo-a5}
\end{figure*}
}

\def\tabExtraVisTwo#1{
\captionsetup[sub]{font=small}
\begin{figure*}[#1]
  \vspace{15pt}
  \centering
  \includegraphics[width=0.95\linewidth]{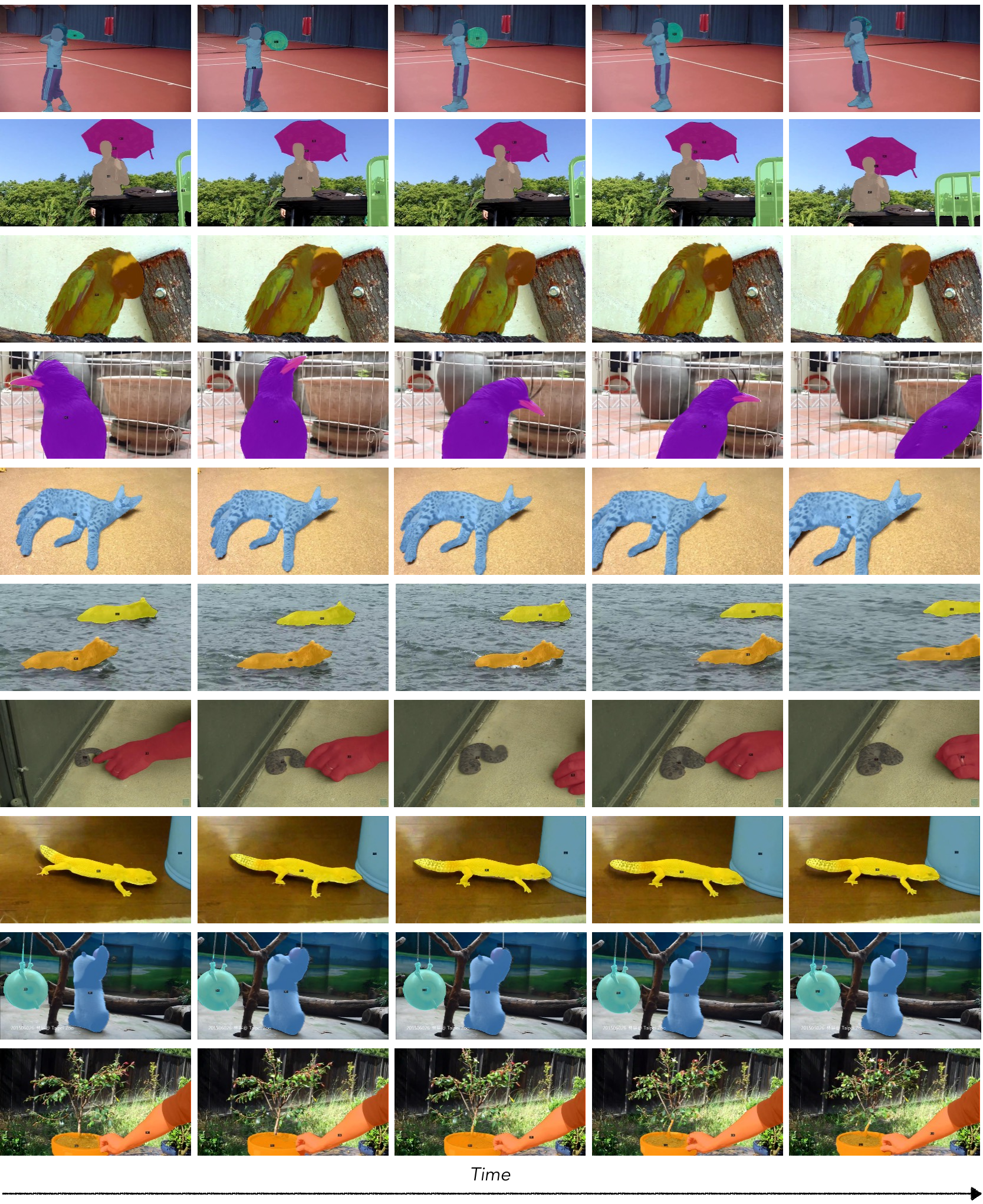}
  \caption{
    Additional qualitative visualizations.
  }
  \label{fig:demo-a6}
\end{figure*}
}

%%%%%%%%%%%%%%%%%%%%%%%%%%%%%%%%%%%%%%%%%%%%%%%%%
% \clearpage

\appendix
\section{Appendix}
\counterwithin{figure}{section}
\setcounter{figure}{0}

\tabExtraVis{!ht}

\tabExtraVisFive{h}
\tabExtraVisFour{h}
\tabExtraVisThree{h}
\tabExtraVisTwo{h}

\label{appendix:training-details}
% \clearpage

% WARNING: do not forget to delete the supplementary pages from your submission 
% \input{sec/X_suppl}

\end{document}